\documentclass[10pt,twocolumn,letterpaper]{article}

\usepackage{iccv}
\usepackage{times}
\usepackage{epsfig}
\usepackage{graphicx}
\usepackage{amsmath}
\usepackage{amssymb}
\usepackage{multirow}
\usepackage{booktabs}
\usepackage{float}
\usepackage{caption}
\usepackage{subcaption}
\usepackage{soul}
\usepackage{arydshln}
\usepackage[accsupp]{axessibility}

\usepackage[pagebackref=true,breaklinks=true,letterpaper=true,colorlinks,bookmarks=false]{hyperref}

\iccvfinalcopy 

\def\iccvPaperID{9596} 

\ificcvfinal\fi

\begin{document}

\title{LFS-GAN: Lifelong Few-Shot Image Generation}

\author{Juwon Seo$^{1}$\thanks{Equal contribution}
\and
Ji-Su Kang$^{2*}$
\and
Gyeong-Moon Park$^{1}$\thanks{Corresponding author}
\and
\\
$^1$Kyung Hee University, Yongin, Republic of Korea \\
$^2$KLleon Tech., Seoul, Republic of Korea
\and
\tt{\small{\{jwseo001, gmpark\}@khu.ac.kr jisu.kang@klleon.io}}
}
\maketitle
\ificcvfinal\thispagestyle{empty}\fi

\begin{abstract}
We address a challenging lifelong few-shot image generation task for the first time.
In this situation, a generative model learns a sequence of tasks using only a few samples per task.
Consequently, the learned model encounters both catastrophic forgetting and overfitting problems at a time.
Existing studies on lifelong GANs have proposed modulation-based methods to prevent catastrophic forgetting.
However, they require considerable additional parameters and cannot generate high-fidelity and diverse images from limited data.
On the other hand, the existing few-shot GANs suffer from severe catastrophic forgetting when learning multiple tasks.
To alleviate these issues, we propose a framework called Lifelong Few-Shot GAN (LFS-GAN) that can generate high-quality and diverse images in lifelong few-shot image generation task.
Our proposed framework learns each task using an efficient task-specific modulator - Learnable Factorized Tensor (LeFT).
LeFT is rank-constrained and has a rich representation ability due to its unique reconstruction technique.
Furthermore, we propose a novel mode seeking loss to improve the diversity of our model in low-data circumstances.
Extensive experiments demonstrate that the proposed LFS-GAN can generate high-fidelity and diverse images without any forgetting and mode collapse in various domains, achieving state-of-the-art in lifelong few-shot image generation task.
Surprisingly, we find that our LFS-GAN even outperforms the existing few-shot GANs in the few-shot image generation task.
The code is available at \href{https://github.com/JJuOn/LFS-GAN}{Github}.
\end{abstract}
\begin{figure*}[!]
    \begin{center}
        \includegraphics[width=0.8\linewidth]{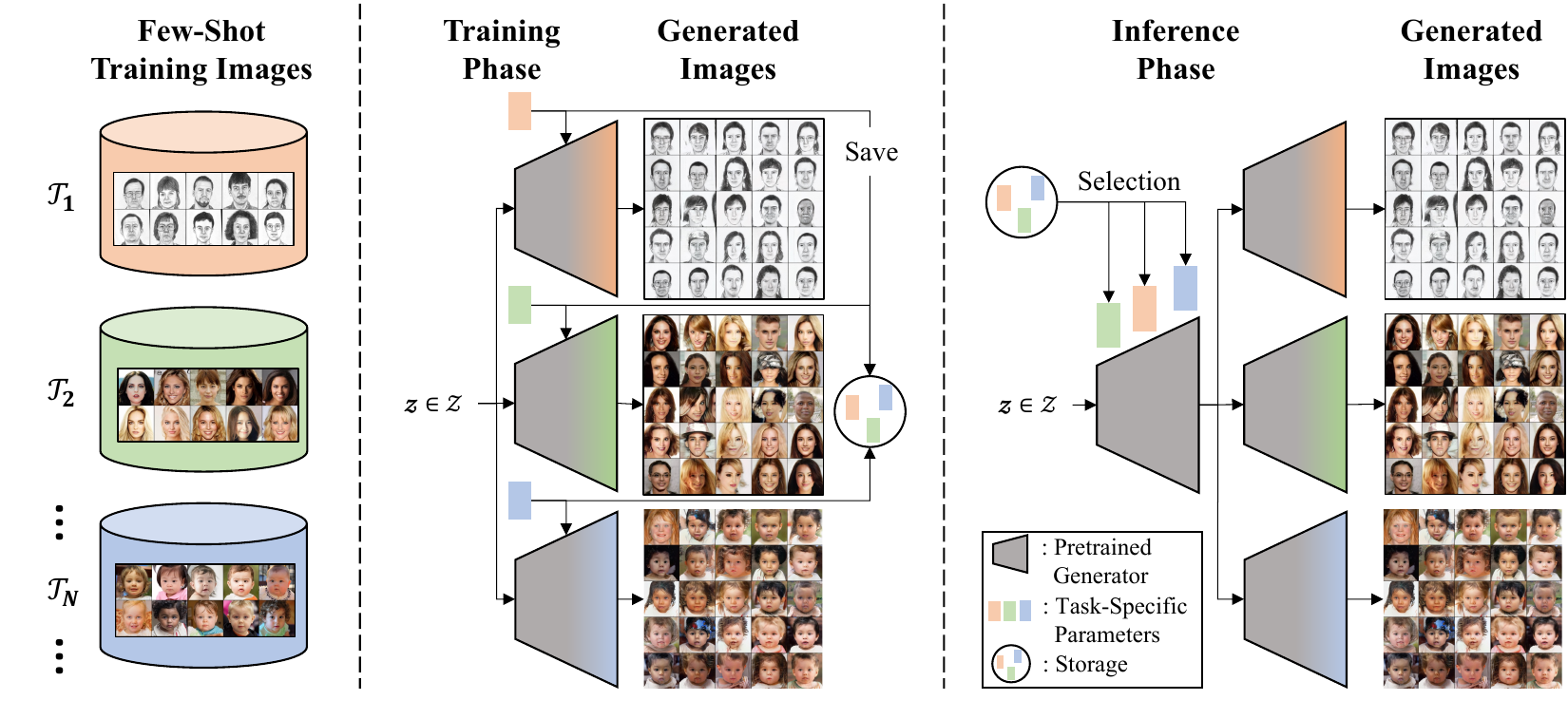}
    \end{center}
    \caption{Illustration of our proposed lifelong few-shot image generation task. We construct a sequence of few-shot tasks. In the training phase, the model learns each task from the pretrained model. By updating only task-specific parameters applied to the model, the model can learn the current task without forgetting. In the inference phase, we can generate high-fidelity and diverse images of not only the current task but also the previous tasks by adopting task-specific parameters to the model.}
    \label{fig:1}
\end{figure*}
\vspace{-0.5cm}
\section{Introduction}
Deep learning has achieved remarkable success in recent years, particularly in a single task learning on a large dataset such as ImageNet \cite{deng2002imagenet} or FFHQ \cite{Karras_2019_CVPR}.
However, obtaining a large amount of refined data for real-world applications is prohibitively expensive, and there are many domains where only limited data can be collected, such as the artistic domain.
Additionally, when faced with long sequences of tasks over time, it is inevitable to train a new model for each target task.

In this context, many studies have recently highlighted the importance of lifelong few-shot learning.
Lifelong few-shot learning combines two challenging settings: lifelong learning and few-shot learning.
It seeks to overcome catastrophic forgetting \cite{MCCLOSKEY1989109, FRENCH1999128, kirkpatrick2017overcoming} when learning a sequence of tasks over time, while learning from limited data without any overfitting problem.
By integrating the concepts of lifelong and few-shot learning, lifelong few-shot learning holds great potential for real-world applications where data is limited or costly to obtain, and where learning a new model for each task is not practical.

Previous studies on lifelong few-shot learning have primarily focused on discriminative tasks \cite{akyurek2022subspace,mazumder2021few,jin2021learn,shi2021overcoming,qin2022lfpt,tao2020few}.
However, lifelong few-shot learning on generative tasks is unexplored before.
Lifelong few-shot image generation task involves training a model to generate realistic and diverse images from handful training images, while continually learning new tasks and preserving the ability to generate images from the previous domains (see Figure \ref{fig:1}).
There are two key challenges in this setting.
First, since the model learns tasks sequentially, it easily forgets the ability to generate samples of the previous tasks.
Second, since the model learns from a biased and sparse distribution, it suffers from the mode collapse problem \cite{arjovsky2017wasserstein}, i.e., it is prone to re-generate the same training samples or produce similar images regardless of the noises provided.
Addressing these challenges is critical to enable the successful application of lifelong few-shot image generation in the real-world, where data is limited or costly to obtain.

To alleviate catastrophic forgetting, recent studies in the field of lifelong image generation have proposed a weight modulator inspired by the affine transformation \cite{huang2017arbitrary}, which enables generative models to learn task-specific information in lifelong setting \cite{zhai2021hyper, cong2020gan, verma2021efficient, varshney2021cam}.
However, conventional generative models \cite{Karrs2018progressive,Karras_2019_CVPR,karras2020analyzing, Karras2021} consist of convolutional layers with high-dimensional weights, resulting in a significant increase of the number of parameters required to modulate them.
As the number of tasks increases, the memory required to store these parameters becomes a significant burden.
Furthermore, since lifelong generative models are designed to synthesize decent images from sufficient data, they suffer from mode collapse with handful training images.
In the realm of few-shot image generation task, many works propose regularization-based methods to maintain the rich diversity of source models \cite{li2020few, ojha2021few, xiao2022few, zhao2022closer}.
However, they are prone to forget how to generate the previous tasks when learning a new task, since they fine-tune the models to learn the new task.
A recent study \cite{zhao2022fewshot} has introduced a modulation-based approach. This method divides the weights into be modulated and fine-tuned parts, based on their significance. Although this work adopts a modulation technique, it suffers from catastrophic forgetting due to its fine-tuning part.

To address these challenges, we propose a novel framework Lifelong Few-Shot Generative Adversarial Network (LFS-GAN).
Our LFS-GAN learns a new task via a powerful weight modulation technique called the Learnable Factorized Tensor (LeFT) that captures task-specific knowledge with low memory costs while freezing learned weights from the source task.
Our proposed LeFT reduces the memory burden by decomposing the weight tensor and restoring it during the forward operation to modulate the weight.
This method enables the efficient and effective generation of high-quality images.
Furthermore, we propose a cluster-wise mode seeking loss to improve the diversity of generated images.
The mode seeking loss \cite{mao2019mode} has shown its effect to diversify the generated images of GANs.
However, in a low-data circumstance, a simple application of mode seeking loss shows less effect because the generated images tend to be similar to the training images.
Thus, we alter the mode seeking loss to be effective in our task and achieve greater diversity.
Lastly, we find that the intra-cluster LPIPS cannot capture the imbalanced generation with respect to training images. To resolve this issue, we propose a novel diversity measure called Balanced Inter- and Intra-cluster LPIPS (B-LPIPS) to accurately evaluate generation diversity in our task.


Our main contributions can be summarized as follows:
\begin{itemize}
    \item To the best of our knowledge, we formulate and tackle a challenging lifelong few-shot image generation task for the first time.
    \item We introduce a novel weight modulation technique, called Learnable Factorized Tensor (LeFT), which enables the generative model to learn new tasks without forgetting and significant parameter growth.
    \item To enhance diversity in the generated images, we propose a cluster-wise mode seeking loss that maximizes the relative distances of intermediate latent codes, feature maps, and images.
    \item Extensive experiments, including our novel metric B-LPIPS, demonstrate that LFS-GAN outperforms the current state-of-the-art methods on generating high-quality and diverse images not only in lifelong few-shot image generation task but also in few-shot image generation task.
    
\end{itemize}

\section{Related Work}
\paragraph{Lifelong Image Generation.}
Recently, several methods have been proposed to alleviate catastrophic forgetting in generative models that are trained on a sequence of tasks continuously.
One such approach is Lifelong GAN \cite{zhai2019lifelong}, which uses a knowledge distillation \cite{hinton2015distilling} based technique to prevent catastrophic forgetting.
As a result, even when the generative model learns a new task, it minimally loses the previously acquired knowledge.
Another method, MeRGAN \cite{wu2018memory}, prevents catastrophic forgetting by utilizing memory replay techniques, such as joint retraining and aligning replays.
During joint retraining, replayed samples are utilized, while the aligning replay forces the current generator to generate the same samples as the auxiliary generator.
GAN-Memory \cite{cong2020gan} proposes a non-forgetting lifelong image generation algorithm by using additional memory to learn the newly arrived task.
It introduces variants of modulation algorithms, such as FiLM \cite{perez2018film} and AdaFM \cite{zhao2020leveraging}, to learn the current task.
Despite the effectiveness of these methods, they have significant limitations in the low-data regime, such as severe mode collapse. 
Additionally, these methods require a large number of parameters for each new task, resulting in accelerating overfitting of the network.
In contrast, our proposed approach takes into consideration the scarcity of training data in few-shot learning scenarios and employs a significantly reduced number of trainable parameters, effectively mitigating the risk of over-fitting.
LoRA \cite{hu2022lora} also studies the efficient fine-tuning technique of Transformers \cite{brown2020language} in natural language processing area.
However, its decomposition and reconstruction schemes are too simple to apply to convolution layers of generative models.
\vspace{-0.5cm}
\paragraph{Few-Shot Image Generation.}
Recently, there has been significant progress in few-shot classification tasks \cite{finn2017model,baik2020meta,baik2021meta, Hu2022pushing}.
This has led to a great interest in few-shot image generation, both in conditional and unconditional settings.
The goal of few-shot image generation task is to generate realistic and diverse images from a limited number of training samples.
FUNIT \cite{li2020few} explores image-to-image translation between source and target domains in few-shot context.
Fusion-based methods, such as F2GAN \cite{hong2020f2gan}, LoFGAN \cite{Gu2021lofgan}, and AGE \cite{ding2022attribute}, have also studied few-shot image generation in conditional settings, where there are a fixed number of training images per class or category.
Our work focuses on a more challenging setting, where we have only a limited number of images per dataset or domain.
Other studies on the unconditional setting are FastGAN \cite{liu2021towards}, which proposes a fast and stabilized GAN architecture and a self-supervised learning method, and MoCA \cite{li2022prototype}, which employs a prototype memory with an attention mechanism \cite{vaswani2017attention}.
However, these studies tackle scenarios where the number of training samples is more than a hundred, while our approach addresses the extreme few-shot setting, where only ten training samples are available.

Many recent works have shown that leveraging pre-trained networks trained on large datasets can be effective in the extreme low-data regime.
For instance, TGAN \cite{wang2018transferring} argues that fine-tuning a network from a large source network can lead to effective results in few-shot setting.
FreezeD \cite{mo2020freeze} fine-tunes a pre-trained GAN by freezing the earlier layers of the discriminator.
BSA \cite{noguchi2019image} fine-tunes the pre-trained network by adapting batch statistics, while EWC \cite{li2020few} uses the Fisher information matrix to prevent changes in important weights.
CDC \cite{ojha2021few} preserves pairwise distances among generated samples, and RSSA \cite{xiao2022few} utilizes a self-correlation matrix for structural consistency.
DCL \cite{zhao2022closer} maximizes mutual information using contrastive loss \cite{chen2020simple}.

Recent work, AdAM \cite{zhao2022fewshot}, adopts a modulation-based approach in few-shot image generation, splitting the weights into modulated and fine-tuned components to generate appropriate images in large domain gaps.
Motivated by AdAM, our proposed approach also employs a modulation-based approach.
Unlike AdAM, our method shows no forgetting in lifelong few-shot image generation and exhibits superior generation quality and diversity.
\section{Method}
In this section, we first formulate lifelong few-shot image generation task (Section \ref{sec:def}).
To learn a generative model for this task, we propose a novel framework - Lifelong Few-Shot GAN (LFS-GAN).
In Section \ref{sec:4-1}, we introduce a lightweight modulation technique - Learnable Factorized Tensor.
To enhance the generation diversity of our LFS-GAN, we adopt a variant of mode seeking loss, described in Section \ref{sec:4-2}.
Furthermore, we point out shortcomings of the existing metric for detecting imbalanced generation and propose a novel diversity measure i.e., Balanced Inter- and Intra-LPIPS (B-LPIPS) in Section \ref{sec:4-3}.

\subsection{Lifelong Few-Shot Image Generation Task}
\label{sec:def}
In this section, we first define lifelong few-shot image generation task.
As illustrated in Figure \ref{fig:1}, given a sequence of tasks \(T=\{\mathcal{T}_1,\mathcal{T}_2,...,\mathcal{T}_N\}\), each task consists of a dataset which contains the images, denoted as \(\mathcal{D}^t=\{x^t_i\}^k_{i=1}\).
Here, \(k\) means the number of training samples, so we can call our task as \(k\)-shot image generation task.
With few training samples, the model easily converges to the biased distribution, which causes the model to be overfitted.
Moreover, as the given sequence of tasks becomes longer, the trained model significantly forgets the previous tasks.
Our goal is to train a model to have the following ability: after training on the \(t^{th}\) task, the trained model can generate the realistic and diverse samples of both the current task \(\mathcal{T}_t\) and the previous tasks \(\{\mathcal{T}_1,...,\mathcal{T}_{t-1}\}\).

\subsection{Learnable Factorized Tensor}
\label{sec:4-1}
\begin{figure}[!]
    \begin{center}
    \includegraphics[width=\linewidth]{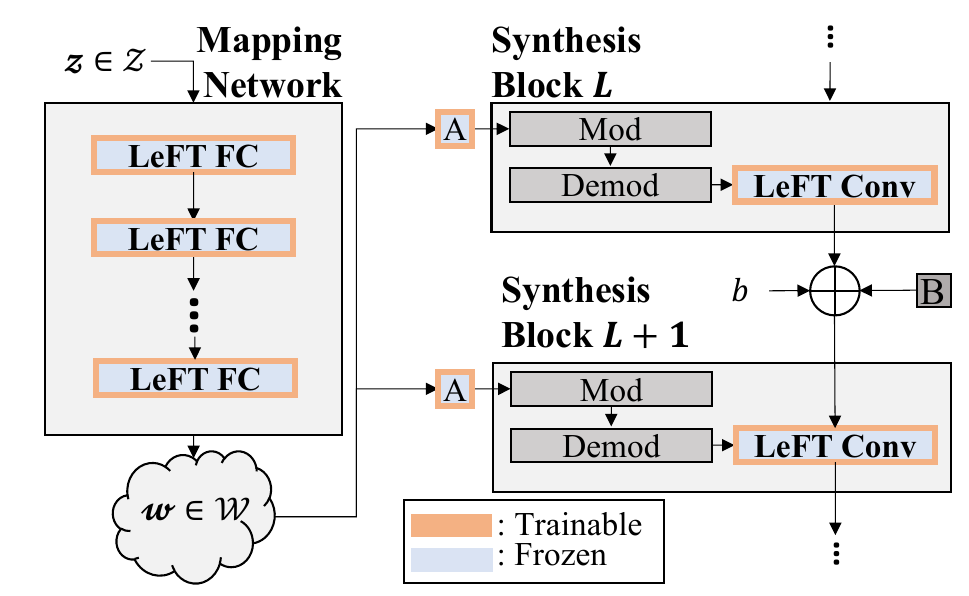}
    \end{center}
    \caption{StyleGAN2 generator architecture with our proposed Learnable Factorized Tensor (LeFT). "A" and "B" denote the affine transformation and noise injection used in the original StyleGAN2 implementation, respectively. We apply LeFT on FC layers and conv. layers to modulate the original weights. The original weights are kept frozen during learning a new task while only LeFT is trainable.}
    \label{fig:2}
\end{figure}
To train a generative model on a sequence of tasks without catastrophic forgetting on the previous tasks, we adopt a modulation-based approach.
Inspired by the style transfer literature \cite{huang2017arbitrary}, we modulate the pretrained weights by affine transformation.
The affine transformation consists of two operations - multiplication and addition.
We use StyleGAN2 \cite{karras2020analyzing} as our backbone, and each convolution layer has a weight tensor $\mathrm{\mathbf{W}}\in\mathbb{R}^{c_{out} \times c_{in} \times k \times k}$.
Here, $c_{out}$ and $c_{in}$ denote the size of the output and input channels, respectively, and $k$ is the kernel size.
We modulate the original weight tensor $\mathrm{\mathbf{W}}$ to obtain the modulated weight tensor $\hat{\mathrm{\mathbf{W}}}\in\mathbb{R}^{c_{out}{\times}c_{in}{\times}k{\times}k}$ like:
\begin{align}
\hat{\mathrm{\mathbf{W}}}=\mathrm{\mathbf{W}} \odot \mathrm{\mathbf{\Gamma}} + \mathrm{\mathbf{B}},
\end{align}
where $\mathrm{\mathbf{\Gamma}}\in\mathbb{R}^{c_{out} \times c_{in} \times k \times k}$ and $\mathrm{\mathbf{B}}\in\mathbb{R}^{c_{out} \times c_{in} \times k \times k}$ are the task-specific modulation parameters which are responsible for multiplication and addition, respectively, and $\odot$ is a Hadamard-product.
In training on each task, we set the original weight $\mathrm{\mathbf{W}}$ frozen and only the modulation parameters $\mathrm{\mathbf{\Gamma}},\mathrm{\mathbf{B}}$ trainable.
By training only the modulation parameters, we can learn new tasks while not updating the pre-trained weights.
Therefore, the model can learn multiple tasks without forgetting.
We also apply the same approach to the fully-connected layers which consist of the weight tensor $\mathrm{\mathbf{W}}_{FC}\in\mathbb{R}^{d_{out}{\times}d_{in}}$, where $d_{out}$ and $d_{in}$
are the dimensions of the output and input, respectively.
In Figure \ref{fig:2}, we demonstrate which layers of the StyleGAN2 generator are modulated.
\begin{figure}[!]
    \includegraphics[width=\linewidth]{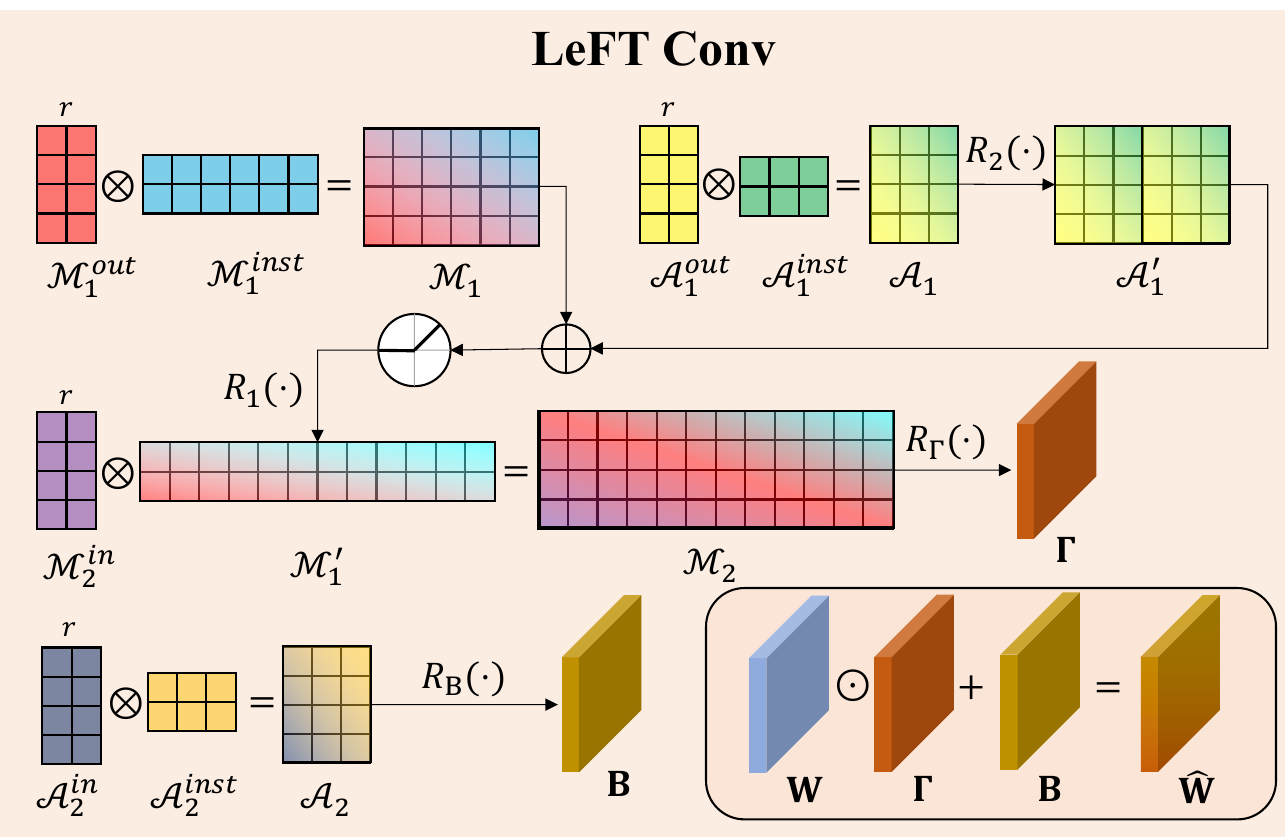}
    \caption{The reconstruction process of LeFT. The above example assumes that the rank $r$=2.}
    \label{fig:3}
\end{figure}
However, as the weight tensor of convolution layers is high-dimensional, storing task-specific modulation parameters for each task results in a serious memory burden.
Furthermore, it is well-known that adapting a number of parameters to learn each low-shot task accelerates the over-fitting problem.
To address these issues, we propose a novel weight decomposition technique called Learnable Factorized Tensor (LeFT).
A key operation of LeFT is a rank-constrained decomposition.
For example, we first reshape one of the modulation parameter $\mathrm{\mathbf{\Gamma}}$ to be three-dimensional $\mathrm{\mathbf{\Gamma}}\in\mathbb{R}^{c_{out}\times{c_{in}}\times{K}}$, where $K$ is a square of kernel size.
Therefore, the original tensor can be expressed using three two-dimensional matrices: $\mathcal{M}^{out}_1 \in \mathbb{R}^{c_{out} \times r}, \mathcal{M}^{inst}_1 \in \mathbb{R}^{r \times r \cdot K}$, and $\mathcal{M}^{in}_2 \in \mathbb{R}^{c_{in} \times r} $, where $r$ is a rank.
We can reconstruct the original tensor by:
\begin{align}
    \label{eq:2}
   \mathcal{M}_1 &= \mathcal{M}^{out}_1 \otimes
   \mathcal{M}^{inst}_1,\\
   \label{eq:3}
   \mathcal{M}_1' &= {R_1}(\mathcal{M}_1),\\
   \mathcal{M}_2 &= \mathcal{M}^{in}_2 \otimes \mathcal{M}_1',\\
   \mathrm{\mathbf{\Gamma}} &= {R_\Gamma}(\mathcal{M}_2),
\end{align}
where $\otimes$ is a matrix-multiplication operation, and ${R_1}\colon\mathbb{R}^{c_{out} \times r \cdot K} \to \mathbb{R}^{r \times c_{out} \cdot K}$ and ${R_\Gamma}\colon\mathbb{R}^{c_{in} \times c_{out} \cdot K} \to \mathbb{R}^{c_{out} \times c_{in} \times k \times k}$ are the reshaping functions.
For the addition parameter $\mathrm{\mathbf{B}}$, we reconstruct by using two-dimensional matrices: $\mathcal{A}^{in}_2 \in \mathbb{R}^{c_{in} \times r}$ and $\mathcal{A}^{inst}_2 \in \mathbb{R}^{r \times K}$:
\begin{align}
    \mathcal{A}_2 &= \mathcal{A}^{in}_2 \otimes \mathcal{A}^{inst}_2,\\
    \mathrm{\mathbf{B}} &= {R_B}(\mathcal{A}_2),
\end{align}
where $R_B\colon \mathbb{R}^{c_{in} \times K} \to \mathbb{R}^{c_{out} \times c_{in} \times k \times k}$ is a function operating repetition and reshaping.
By the above efficient decomposition and reconstruction scheme, we can reduce the number of modulation parameters significantly about less than 1\%.

From Equation \ref{eq:2}, recovering a three-dimensional tensor through multiplication is similar to the operation of multi-layered perceptron (MLP).
In MLP, using bias and activation functions can generally improve performance.
To apply bias and activation to the LeFT, we additionally introduce two matrices: $\mathcal{A}^{out}_1\in\mathbb{R}^{c_{out}\times r}$ and $\mathcal{A}^{inst}_1\in\mathbb{R}^{r\times K}$.
These two matrices are also rank-constrained.
We can reconstruct the bias $\mathcal{A}'_1 \in \mathbb{R}^{c_{out} \times r \cdot K}$ as:
\begin{align}
    \mathcal{A}_1 &= \mathcal{A}^{out}_1 \otimes \mathcal{A}^{inst}_1, \\
    \mathcal{A}_1' &= R_2(\mathcal{A}_1),
\end{align}
where $R_2\colon \mathbb{R}^{c_{out} \times K} \to \mathbb{R}^{c_{out} \times r \cdot K}$ is a repetition function.
Instead of Equation \ref{eq:3}, we add the bias $\mathcal{A}_1$ to the intermediate matrix $\mathcal{M}_1$ and apply activation function $\mathrm{act(\cdot)}$ as follows:
\begin{align}
    \mathcal{M}_1' = R_1(\mathrm{act}(\mathcal{M}_1 + \mathcal{A}_1')).
\end{align}
We experimentally find that using ReLU function as the activation function is most effective. The overall process is shown in Figure \ref{fig:3}.
We also apply LeFT on fully-connected layers.
The decomposition and reconstruction processes are described in the Supplementary.
\subsection{Cluster-Wise Mode Seeking Loss}
\label{sec:4-2}
Diverse image generation is a major interest when the number of training images is extremely small.
Due to the lack of concerns on diversity, the existing modulation-based methods \cite{cong2020gan, varshney2021cam, zhao2022fewshot} show inferior diversity performance compared to regularization-based \cite{li2020few, ojha2021few, xiao2022few, zhao2022closer} methods.
Furthermore, rigid regularizations to preserve the diversity of source domain often result in unnatural distortions on generated images.
To enhance diversity within the target domain, the mode seeking loss \cite{mao2019mode} and its variants \cite{cheng2022inout} have shown diverse image generation.
The mode seeking loss introduced in \cite{mao2019mode} is as follows:
\begin{align}
    \mathcal{L}_{ms} = \min({\frac{\Delta{z}}{\Delta{I}}}),
\end{align}
where $I$ and $z$ are the generated images and the input noise vectors, respectively, and $\Delta$ is a mean absolute error as a distance measure.
The mode seeking loss maximizes the distance of images with respect to the distance of the input noise vectors.
However, in few-shot setting, the generated images are prone to be close to the given training images.
In this situation, applying the original mode seeking loss is not effective.
Therefore, we propose a cluster-wise mode seeking loss - a variant of mode seeking loss which is effective in few-shot scenario.
Initially, we set clusters as many as $B$, a batch size of the training images.
We generate $4B$ images and assign each of them to the perceptually closest cluster.
Different from \cite{mao2019mode}, we utilize the intermediate latent vector $w\in\mathcal{W}$ and the feature map $F_l$ of layer $l$.
We maximize $w_i$, $F_{l,i}$, and $I_i$ with respect to $z_i$, $w_i$, and $w_i$ within each cluster $c_i$.
Thus, our cluster-wise mode seeking loss is computed as:
\begin{align}
    d_{w,i} &= \frac{\Delta w_i}{\Delta z_i}, ~ d_{F,i}= \frac{1}{L}\sum^L_{l=1}\frac{\Delta F_{l,i}}{\Delta w_i}, ~ d_{I,i} = \frac{\Delta I_i}{\Delta w_i}, \\
    \mathcal{L}_{cms} &= \min(\cfrac{1}{\frac{1}{B}{\sum^B_{i=1}}\left[ d_{w,i} + d_{F,i} + d_{I,i} \right]}).
\end{align}
We apply this cluster-wise mode seeking loss to update the generator.
The total loss functions for the generator and the discriminator the are:
\begin{align}
    \mathcal{L}^G &= \mathcal{L}^G_{adv} + \lambda\mathcal{L}_{cms}, \\
    \mathcal{L}^D &= \mathcal{L}^D_{adv},
\end{align}
where $\mathcal{L}^G_{adv}$ and $\mathcal{L}^D_{adv}$ are the non-saturating adversarial losses proposed in \cite{goodfellow2014generative}, and $\lambda$ is a hyper-parameter that controls the effect of the proposed cluster-wise mode seeking loss.
We experimentally find that using $\lambda=1$ is the most effective.

\begin{figure}[!]
    \begin{subfigure}[t]{0.49\columnwidth}
        \includegraphics[width=\textwidth]{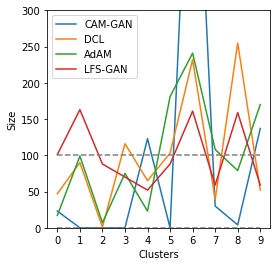}
        \caption{The size of each cluster.}
    \end{subfigure}
    \hfill
    \begin{subfigure}[t]{0.49\columnwidth}
        \includegraphics[width=\textwidth]{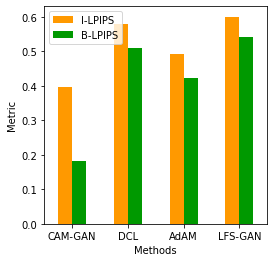}
        \caption{I-LPIPS vs. B-LPIPS.}
    \end{subfigure}
    \caption{We measure the number of each cluster and its corresponding I-LPIPS and  B-LPIPS on learning Babies task.
    The number of fake image is 1000 and the gray line represents the ideal number 100. We observe (a): the distribution of each cluster's size is highly imbalanced, (b): the I-LPIPS cannot reflect this biased distribution.}
    \label{fig:4}
\end{figure}

\begin{figure*}[!]
    \begin{center}
    \resizebox{!}{0.4\textheight}{
    \includegraphics{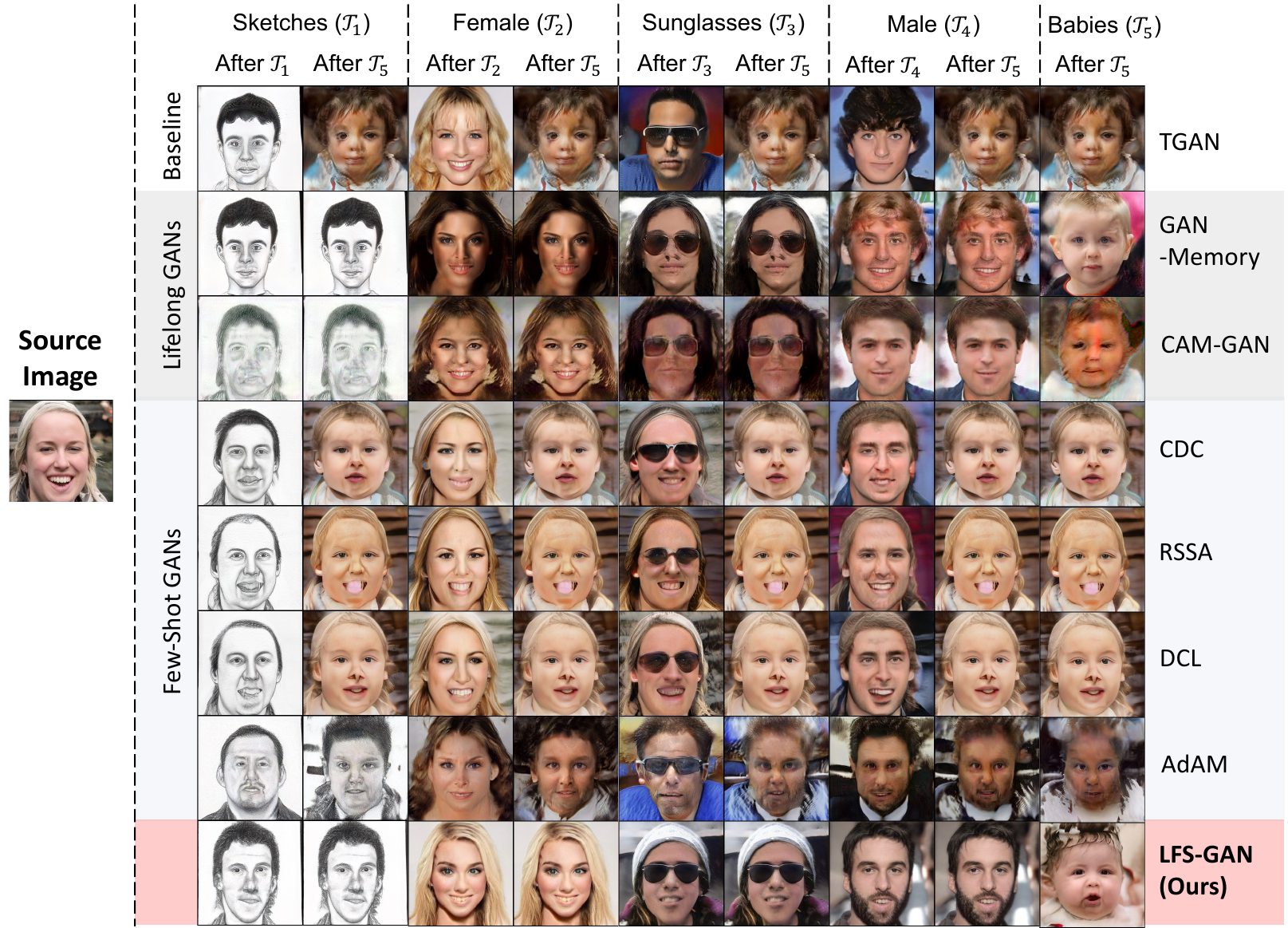}
    }
    \end{center}
    \caption{Qualitative comparison on lifelong few-shot image generation task. Given source image of left, we visualize the generated image of after each task and after the last task.}
    \vspace{-0.3cm}
    \label{fig:6}
\end{figure*}
\subsection{Balanced Inter- and Intra-Cluster LPIPS}
\label{sec:4-3}

Introduced in \cite{ojha2021few}, intra-cluster LPIPS (I-LPIPS) has been widely used to measure diversity in few-shot image generation task \cite{zhao2022closer, zhao2022fewshot}.
In I-LPIPS metric, the fake images are clustered to the real training images by the nearest LPIPS metric \cite{zhang2018unreasonable}.
As a result, I-LPIPS is computed by averaging pairwise LPIPS distance within each cluster.
However, we empirically find that in some cases, the size of the clusters can be highly imbalanced. Furthermore, there are cases where some clusters contain no generated images that are related to the corresponding training images.
In such a situation, intra-cluster LPIPS can not account for this imbalance.
Figure \ref{fig:4} shows an example of the above situation.

To measure diversity in aware of cluster balance, we propose a novel metric called Balanced Inter- and Intra-Cluster LPIPS (B-LPIPS).
We introduce a concept of entropy to reflect the balance of each cluster.
For cluster $c_i$, we can compute the proportion $p(c_i)$ of the cluster:
\begin{align}
    p(c_i) = \frac{\Vert c_i \Vert}{N},
\end{align}
where $\Vert c_i \Vert$ is the size of the cluster $c_i$ and $N$ is the total number of generated images.
The B-LPIPS are computed as a weighted sum of pairwise LPIPS of each cluster:
\begin{align}
    \mathrm{w}_i &= -p(c_i) \cdot \log_{10}{p(c_i)}, \\
    \mathrm{B}\text{-}\mathrm{LPIPS} &= \sum^k_{i=1} \mathrm{w}_i \cdot \mathrm{P}\text{-}\mathrm{LPIPS}(c_i),
\end{align}
where $k$ denotes the number of training images and $\mathrm{P}\text{-}\mathrm{LPIPS}(c_i)$ is a pairwise LPIPS within cluster $c_i$.


\begin{table*}[!]
\begin{center}
\resizebox{0.9\textwidth}{!}{
\begin{tabular}{l|lcccccccccccc}
\toprule[1.1pt]
\multicolumn{2}{l}{\multirow{3}{*}{Method}} &
  \multicolumn{10}{c}{Task Order} &
  \multicolumn{2}{c}{\multirow{2}{*}{Average}} \\ \cline{3-12}
\multicolumn{2}{l}{} &
  \multicolumn{2}{c}{Sketches $(\mathcal{T}_1)$} &
  \multicolumn{2}{c}{Female $(\mathcal{T}_2)$} &
  \multicolumn{2}{c}{Sunglasses $(\mathcal{T}_3)$} &
  \multicolumn{2}{c}{Male $(\mathcal{T}_4)$} &
  \multicolumn{2}{c}{Babies $(\mathcal{T}_5)$} &
  \multicolumn{2}{c}{} \\ \cline{3-14} 
\multicolumn{2}{l}{} &
  FID &
  B-LPIPS &
  FID &
  B-LPIPS &
  FID &
  B-LPIPS &
  FID &
  B-LPIPS &
  FID &
  B-LPIPS &
  \multicolumn{1}{c}{FID $(\downarrow)$} &
  B-LPIPS {$(\uparrow)$} \\ \hline
\multicolumn{2}{l}{TGAN (Baseline) \cite{wang2018transferring}} &
  372.89 &
  0.157 &
  255.53 &
  0.238 &
  309.13 &
  0.247 &
  281.43 &
  0.129 &
  171.19 &
  0.203 &
  278.03 &
  0.195 \\ \hline
\multirow{2}{*}{\begin{tabular}[c]{@{}l@{}}Lifelong\\ GANs\end{tabular}} &
  GAN-Memory \cite{cong2020gan}&
  \underline{69.58} &
  \underline{0.311} &
  \underline{71.56} &
  0.287 &
  87.02 &
  0.169 &
  99.44 &
  0.143 &
  177.73 &
  0.150 &
  101.05 &
  0.212 \\
 &
  CAM-GAN \cite{varshney2021cam} &
  91.81 &
  0.293 &
  85.68 &
  0.332 &
  \underline{86.81} &
  0.333 &
  \underline{82.83} &
  0.312 &
  146.20 &
  0.181 &
  \underline{98.66} &
  0.290 \\ \hline
\multirow{4}{*}{\begin{tabular}[c]{@{}l@{}}Few-Shot\\ GANs\end{tabular}} &
  CDC \cite{ojha2021few}&
  322.72 &
  0.205 &
  197.40 &
  0.427 &
  244.94 &
  0.463 &
  227.00 &
  0.381 &
  \underline{69.98} &
  0.454 &
  208.41 &
  0.386 \\
 &
  RSSA \cite{xiao2022few}&
  308.00 &
  0.285 &
  175.20 &
  \underline{0.440} &
  207.58 &
  0.484 &
  205.49 &
  0.405 &
  76.70 &
  0.481 &
  194.59 &
  0.419 \\
 &
  DCL \cite{zhao2022closer}&
  297.73 &
  0.307 &
  170.31 &
  0.435 &
  191.54 &
  \underline{0.490} &
  194.42 &
  \underline{0.443} &
  77.22 &
  \underline{0.487} &
  186.25 &
  \underline{0.432} \\
 &
  AdAM \cite{zhao2022fewshot}&
  161.48 &
  0.250 &
  179.69 &
  0.342 &
  217.99 &
  0.352 &
  163.87 &
  0.299 &
  110.08 &
  0.407 &
  166.82 &
  0.330 \\ \hline
\multicolumn{2}{l}{\textbf{LFS-GAN (Ours)}} &
  \textbf{34.66} &
  \textbf{0.354} &
  \textbf{29.59} &
  \textbf{0.481} &
  \textbf{27.69} &
  \textbf{0.584} &
  \textbf{35.44} &
  \textbf{0.472} &
  \textbf{41.48} &
  \textbf{0.556} &
  \multicolumn{1}{c}{\textbf{33.77}} &
  \textbf{0.489} \\ \bottomrule[1.1pt]
\end{tabular}
}
\end{center}
\caption{Quantitative results on \textbf{lifelong few-shot image generation} task. We measured each metric after the last task - Babies ($\mathcal{T}_5$). The bold value represents the best result and the underlined value represents the second best result.}
\label{table:1}
\end{table*}
\section{Experiments}
\subsection{Exeperimental Setup}
\paragraph{Datasets.}
We used FFHQ \cite{Karras_2019_CVPR}, LSUN-Church, and LSUN-Cars \cite{yu2015lsun} as the source domains.
The target domains were (i) Sketches \cite{ojha2021few}, (ii) Female \cite{Karrs2018progressive}, (iii) Sunglasses \cite{ojha2021few}, (iv) Male \cite{Karrs2018progressive}, (v) Babies \cite{ojha2021few}, (vi) Van Gogh's house (vii) Haunted house, (viii) Palace, and (ix) Abandoned cars.
For source domains of LSUN-Church and LSUN-Cars, we presented experimental results in the Supplementary.
\vspace{-0.5cm}
\paragraph{Baselines.}
Since our work tackled lifelong few-shot for the first time.
We established our work on the baseline - (i) TGAN \cite{wang2018transferring}.
We evaluated two distinct approaches to our proposed lifelong few-shot image generation task - (1) lifelong GANs and (2) few-shot GANs.
For lifelong GANs, we considered two methods: (ii) GAN-Memory \cite{cong2020gan} and (iii) CAM-GAN \cite{varshney2021cam} are modulation-based lifelong GANs.
For few-shot GANs, we evaluated four methods: (iv) CDC \cite{ojha2021few}, (v) RSSA \cite{xiao2022few}, and (vi) DCL \cite{zhao2022closer} are regularization-based few-shot GANs and (vii) AdAM \cite{zhao2022fewshot} is a modulation-based few-shot GAN.
\vspace{-0.5cm}
\paragraph{Metrics.}
In the evaluation of the generation performance, we adopted three metrics.
Firstly, the Fréchet Inception Distance (FID) score \cite{heusel2017gans} was used to measure the similarity between the generated images and real images.
A lower FID score indicates a higher quality of generation.
Secondly, the balanced inter- and intra-cluster LPIPS (B-LPIPS), which is proposed in our paper, was used to measure the diversity of the generated images.
Finally, we used the intra-cluster LPIPS (I-LPIPS) \cite{ojha2021few}, as a traditional metric for the auxiliary measure of generation diversity.
A higher B-LPIPS or I-LPIPS demonstrates a greater diversity of the generated images, as they are more distinct from one another.
We sampled 5,000 images to compute FID score and 1,000 images to compute both B-LPIPS and I-LPIPS.
In the main paper, we chose B-LPIPS as a default metric for measuring diversity.
The diversity comparsion using I-LPIPS is presented in the Supplementary.
\subsection{Results on LFS Task}
\paragraph{Qualitative Results.}
At first, we evaluated state-of-the-art methods on our proposed lifelong few-shot image generation task.
Figure \ref{fig:6} shows a qualitative result of state-of-the-art methods and LFS-GAN.
As seen on the figure, TGAN and other few-shot GANs suffered from catastrophic forgetting.
Moreover, they showed degraded quality in the last task.
It was because they learned a single model on a sequence of task while ruining the model due to the biased and scarce distribution of each task.
On the other hand, the images from lifelong GANs had a lot of distortions and were similar to the training images.
This result demonstrated that lifelong GANs suffered from mode collapse.
Compared to other methods, ours could generate high-quality and diverse images without forgetting.
\vspace{-0.5cm}
\paragraph{Quantitative Results.}
As seen on Table \ref{table:1}, TGAN could not generate neither high-quality nor diverse images in our task.
Lifelong GANs could generate high-quality images compared to TGAN.
However, they could not generate diverse images.
It was because they had no concerns about the low-data circumstance, thus they suffered from mode collapse.
On the other hand, few-shot GANs generated diverse images, but they failed to generate high-quality images.
We argue that this phenomenon was because of severe catastrophic forgetting happening during learning on a sequence of tasks.
Unlike other methods, AdAM showed the alleviated forgetting.
It is due to its training scheme of separating the weights into be fine-tuned and modulated, thus it could recover some amount of the knowledge of the previous tasks.
However, our LFS-GAN could generate both high-quality and diverse images.
Furthermore, we evaluated state-of-the-art methods on training efficiency.
In Table \ref{table:2}, we find that our LFS-GAN achieved the most efficient parameter consumption to learn a new task.
\begin{table}[h]
\begin{center}
\resizebox{0.45\textwidth}{!}{
\begin{tabular}{llcc}
\toprule[1.1pt]
\multicolumn{2}{l}{Method} & \begin{tabular}[c]{@{}c@{}}\# of\\ Trainable Params.\end{tabular} & \begin{tabular}[c]{@{}c@{}}\% w.r.t.\\ Backbone\end{tabular} \\ \hline
\multicolumn{2}{l}{Baseline (TGAN) \cite{wang2018transferring}}                                                                        & 30.0M & 100\%   \\ \hline
\multicolumn{1}{l|}{\multirow{2}{*}{\begin{tabular}[c]{@{}l@{}}Lifelong\\ GANs\end{tabular}}} & GAN-Memory \cite{cong2020gan}& 5.3M  & 17.7\%  \\
\multicolumn{1}{l|}{}                                                                         & CAM-GAN \cite{varshney2021cam}   & 2.3M  & 7.7\%   \\ \hline
\multicolumn{1}{l|}{\multirow{4}{*}{\begin{tabular}[c]{@{}l@{}}Few-Shot\\ GANs\end{tabular}}} & CDC  \cite{ojha2021few}      & 30.0M & 100.0\% \\
\multicolumn{1}{l|}{}                                                                         & RSSA \cite{xiao2022few}      & 30.0M & 100.0\% \\
\multicolumn{1}{l|}{}                                                                         & DCL  \cite{zhao2022closer}      & 30.0M & 100.0\% \\
\multicolumn{1}{l|}{}                                                                         & AdAM \cite{zhao2022fewshot}      & 18.9M & 63.0\%  \\ \hline
\multicolumn{2}{l}{\textbf{LFS-GAN (Ours)}}                                                                & \textbf{0.1M}  & \textbf{0.3}\%   \\ \bottomrule[1.1pt]
\end{tabular}
}
\end{center}

\caption{Comparison on training efficiency. In the first column, there is the number of parameters to learn a new task of each method. Since we use StyleGAN2 as our backbone, the second column represents the percentage with respect to the number of parameters of the StyleGAN2 generator.}
\label{table:2}
\end{table}
\begin{table*}[!]
\begin{center}
\resizebox{0.9\textwidth}{!}{
\begin{tabular}{l|lcccccccccccc}
\toprule[1.1pt]
\multicolumn{2}{l}{\multirow{3}{*}{Method}} &
  \multicolumn{10}{c}{Tasks} &
  \multicolumn{2}{c}{\multirow{2}{*}{Average}} \\ \cline{3-12}
\multicolumn{2}{l}{} &
  \multicolumn{2}{c}{Sketches} &
  \multicolumn{2}{c}{Female} &
  \multicolumn{2}{c}{Sunglasses} &
  \multicolumn{2}{c}{Male} &
  \multicolumn{2}{c}{Babies} &
  \multicolumn{2}{c}{} \\ \cline{3-14} 
\multicolumn{2}{l}{} &
  FID &
  B-LPIPS &
  FID &
  B-LPIPS &
  FID &
  B-LPIPS &
  FID &
  B-LPIPS &
  FID &
  B-LPIPS &
  FID (↓) &
  B-LPIPS(↑) \\ \hline
\multicolumn{2}{l}{Baseline (TGAN)} &
  60.57 &
  0.335 &
  67.63 &
  0.318 &
  72.81 &
  0.391 &
  73.44 &
  0.319 &
  114.29 &
  0.414 &
  77.75 &
  0.355 \\ \hline
\multirow{2}{*}{\begin{tabular}[c]{@{}l@{}}Lifelong\\ GANs\end{tabular}} &
  GAN-Memory &
  69.58 &
  0.311 &
  71.50 &
  0.287 &
  87.02 &
  0.169 &
  99.44 &
  0.143 &
  177.73 &
  0.150 &
  101.05 &
  0.212 \\
 &
  CAM-GAN &
  91.81 &
  0.293 &
  85.68 &
  0.332 &
  86.81 &
  0.333 &
  82.83 &
  0.312 &
  146.20 &
  0.181 &
  98.66 &
  0.290 \\ \hline
\multirow{4}{*}{\begin{tabular}[c]{@{}l@{}}Few-Shot\\ GANs\end{tabular}} &
  CDC &
  49.19 &
  0.237 &
  \underline{31.26} &
  0.450 &
  36.03 &
  0.505 &
  \underline{41.88} &
  0.435 &
  \underline{64.75} &
  0.496 &
  \underline{44.62} &
  0.425 \\
 &
  RSSA &
  56.25 &
  0.251 &
  34.24 &
  0.467 &
  44.01 &
  0.510 &
  44.83 &
  0.434 &
  72.45 &
  0.486 &
  50.36 &
  0.430 \\
 &
  DCL &
  58.60 &
  \underline{0.353} &
  35.19 &
  \underline{0.468} &
  \underline{33.05} &
  \underline{0.517} &
  44.19 &
  \underline{0.436} &
  66.10 &
  \underline{0.508} &
  47.43 &
  \underline{0.456} \\
 &
  AdAM &
  \underline{45.70} &
  0.325 &
  61.79 &
  0.375 &
  45.55 &
  0.392 &
  61.55 &
  0.338 &
  91.13 &
  0.421 &
  61.14 &
  0.370 \\ \hline
\multicolumn{2}{l}{\textbf{LFS-GAN (Ours)}} &
  \textbf{34.66} &
  \textbf{0.354} &
  \textbf{29.59} &
  \textbf{0.481} &
  \textbf{27.69} &
  \textbf{0.584} &
  \textbf{35.44} &
  \textbf{0.472} &
  \textbf{41.48} &
  \textbf{0.556} &
  \textbf{33.77} &
  \textbf{0.489} \\ \bottomrule[1.1pt]
\end{tabular}
}
\end{center}
\caption{Quantitative results on \textbf{few-shot image generation} task. We conducted each task independently. The bold value represents the best result and the underlined value represents the secondary best result.}
\label{table:3}
\end{table*}
\subsection{Results on FS Task}
Different from the proposed lifelong few-shot image generation task, few-shot image generation task aims to generate decent and diverse images on a single target domain consisting of few-shot data.
\vspace{-0.5cm}
\paragraph{Qualitative Results.}
We present the generated samples of state-of-the-art methods and LFS-GAN on the diverse target domains in the Supplementary. As seen on figure, similar to the results of lifelong few-shot image generation task, lifelong GANs could not generate neither high-quality nor diverse images in few-shot image generation task.
On the other hand, few-shot GANs achieved superior performance compared to lifelong GANs.
However, there were several distortions on generated images.
We insist that these distortions came from strong regularizations.
They restrict the ability to learn target domain, resulting in distortion.
\vspace{-0.5cm}
\paragraph{Quantitative Results.}
As seen on Table \ref{table:3}, the existing lifelong GANs failed to generate high-quality or diverse images.
Surprisingly, we find that our LFS-GAN also outperformed the existing few-shot GANs on few-shot image generation task.

\begin{table}[h]
\centering
\resizebox{0.4\textwidth}{!}{
\begin{tabular}{ccccccccccccccc}
\toprule[1.1pt]
\multirow{2}{*}{Bias} &
  \multirow{2}{*}{$r$} &
  \multirow{2}{*}{\begin{tabular}[c]{@{}c@{}}\# of\\ Trainable Params.\end{tabular}} &
  \multicolumn{2}{c}{Average} \\ \cline{4-5} 
 &
   &
   &
  FID ($\downarrow$)&
  B-LPIPS ($\uparrow$) \\ \hline
\multirow{5}{*}{w/} &
  1 &
  108K &
  \textbf{33.77} &
  \textbf{0.489} \\
 &
  2 &
  192K &
  40.22 &
  \underline{0.423} \\
 &
  4 &
  358K &
  43.00 &
  0.402 \\
 &
  8 &
  695K &
  48.27 &
  0.370 \\
 &
  16 &
  1,380K &
  55.20 &
  0.332 \\ \cline{1-5}
\multirow{5}{*}{w/o} &
  1 &
  \textbf{54K} &
  \underline{39.42} &
  0.416 \\
 &
  2 &
  96K &
  39.56 &
  0.410 \\
 &
  4 &
  180K &
  44.40 &
  0.402 \\
 &
  8 &
  350K &
  48.74 &
  0.374 \\
 &
  16 &
  704K &
  52.42 &
  0.326 \\ \bottomrule[1.1pt]
\end{tabular}
}
\caption{Ablation on the bias and the rank of LeFT.}
\label{table:r}
\end{table}
\begin{table}[h]
\begin{center}
\resizebox{0.3\textwidth}{!}{
\begin{tabular}{ccc}
\toprule[1.1pt]
\multirow{2}{*}{Activation} &
  \multicolumn{2}{c}{Average} \\ \cline{2-3} 
 &
  FID $(\downarrow)$ &
  B-LPIPS $(\uparrow)$ \\ \hline
Identity &
  39.87 &
  0.451 \\
Sigmoid &
  37.76 &
  0.417 \\
Tanh &
  38.25 &
  0.440 \\
LeakyReLU &
  \underline{35.42} &
  0.437 \\
GELU &
  37.78 &
  \underline{0.448} \\
SiLU &
  40.28 &
  0.417 \\
\textbf{ReLU} &
  \textbf{33.77} &
  \textbf{0.489} \\ \bottomrule[1.1pt]
\end{tabular}
}
\caption{Ablation on the activation functions of LeFT.}
\label{table:act}
\end{center}
\end{table}


\begin{table}[h]
\begin{center}
    \resizebox{0.4\textwidth}{!}{
    \begin{tabular}{ccccc}
    \toprule[1.1pt]
    \multicolumn{3}{c}{Maximize} & \multicolumn{1}{c}{Average}     \\ \hline
    $\Delta w/\Delta z$      & $\Delta F/\Delta w$     & $\Delta I/\Delta w$                & B-LPIPS $(\uparrow)$       \\ \hline
             &         &                   & 0.423          \\ \hline
             &         & \checkmark                 & \underline{0.436}    \\
             & \checkmark       &                   & 0.426          \\
    \checkmark        &         &             & 0.435          \\ \hline
    \checkmark       & \checkmark       & \checkmark        & \textbf{0.489} \\ \bottomrule[1.1pt]
    \end{tabular}
    }
\end{center}
\caption{Ablation on the maximization target of the cluster-wise mode seeking loss.}
\label{table:cms}
\end{table}
\subsection{Analysis}
\paragraph{Ablation on the LeFT components.}
We first inspected the effect of the bias term and rank of LeFT in Table \ref{table:r}.
In most cases, the bias was responsible for improving both quality and diversity.
We found that while LeFT without bias and of rank of 1 (the sixth row) reduced the number of trainable parameters a lot, the bias term was more crucial for generating high-quality and diverse images.
As seen on Table \ref{table:act}, we observed that applying activation functions on LeFT generally improved the quality of generated samples and we selected to use ReLU as an activation function of LeFT by its decent performance in general.
\vspace{-0.75cm}
\paragraph{Ablation on the cluster-wise mode seeking loss.}
Finally, we experimented on which part to be maximized in cluster-wise mode seeking loss (see Table \ref{table:cms}). In this table, not to apply cluster-wise mode seeking loss (the first row) showed no effect on enhancing diversity compared to the finalized method (the last row). In general, we decided to maximize all parts by cluster-wise mode seeking loss for improved diversity and a slight gain in quality.

\section{Conclusion}
In this paper, we formulate and tackle the challenging lifelong few-shot image generation for the first time.
To generate high-quality and diverse images in our task, we propose a novel framework Lifelong Few-Shot GAN, LFS-GAN for short.
In LFS-GAN, we learn each task by introducing a novel weight modulation technique Learnable Factorized Tensor (LeFT).
When learning each task, we only train LeFT parameters while freezing the original weights, thus we can achieve lifelong few-shot image generation without forgetting.
Moreover, we propose a variant of mode seeking loss - cluster-wise mode seeking loss to enhance the diversity of generated images with less affecting the quality.
Extensive experiments demonstrate that our LFS-GAN achieves state-of-the-art in generating high-quality and diverse images in both lifelong few-shot image generation task and few-shot image generation task.
\vspace{-0.25cm}
\section*{Acknowledgement}
\vspace{-0.25cm}
This work was supported by the National Research Foundation of Korea (NRF) grant funded by the Korea government (MSIT) (No. 2021R1G1A1094379), and in part by MSIT (Ministry of Science and ICT), Korea, under the ITRC (Information Technology Research Center) support program (IITP-2023-RS-2023-00258649) supervised by the IITP (Institute for Information \& Communications Technology Planning \& Evaluation), and in part by the Institute of Information and Communications Technology Planning and Evaluation (IITP) grant funded by the Korea Government (MSIT) (Artificial Intelligence Innovation Hub) under Grant 2021-0-02068, and by Institute of Information \& communications Technology Planning \& Evaluation (IITP) grant funded by the Korea government (MSIT) (No.RS-2022-00155911, Artificial Intelligence Convergence Innovation Human Resources Development (Kyung Hee University)).
\clearpage
\appendix
\section*{Supplementary Material}
\vspace{-0.5cm}
\begin{figure}[h]
    \resizebox{\columnwidth}{!}{
    \centering
    \includegraphics{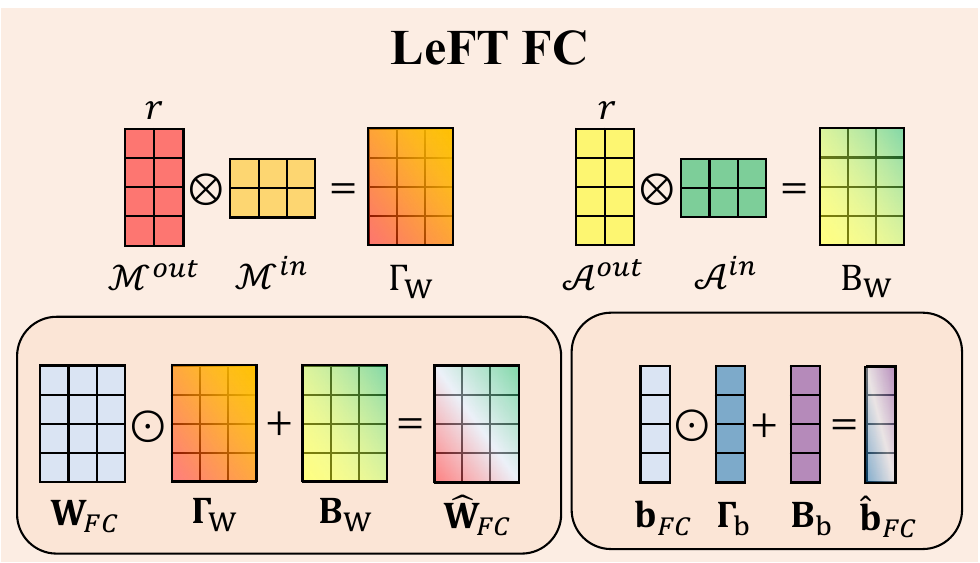}
    }
    \caption{The reconstruction process of LeFT on fully-connected layers. The above example assumes that the rank $r$=2.}
    \label{fig:left_fc}
\end{figure}
\vspace{-0.75cm}
\section{LeFT on Fully-Connected Layers}
Fully-connected layers of StyleGAN2 \cite{karras2020analyzing} architecture consist of weight tensor $\mathbf{W}_{FC} \in \mathbb{R}^{d_{out} \times d_{in}}$ and bias $\mathbf{b}_{FC}\in\mathbb{R}^{d_{out}}$.
We also apply LeFT to fully-connected layers to modulate these parameters like:
\vspace{-0.3cm}
\begin{align}
    \hat{\mathbf{W}}_{FC} &= \mathbf{W}_{FC} \odot \mathbf{\Gamma_W} + \mathbf{B_W}, \\
    \hat{\mathbf{b}}_{FC} &= \mathbf{b}_{FC} \odot \mathbf{\Gamma_b} + \mathbf{B_b},
\end{align}
where $\mathbf{\Gamma_{\{W,b\}}}$ and $\mathbf{B_{\{W,b\}}}$ denote the modulation parameters of LeFT which are responsible for multiplication and addition to modulate weight and bias, respectively.
Dimensions of these modulation parameters are equal to the dimensions of the original parameters - $\mathbf{W}_{FC}$ or ${\mathbf{b}_{FC}}$.
Since $\mathbf{W}_{FC}$ is two-dimensional, we also apply a rank-constrained decomposition similar to the way in the convolution layers described on the main paper.
The overall process is shown on Figure \ref{fig:left_fc}.

Similar to convolutional layers, we decompose $\mathbf{\{\Gamma, B\}_W}$ into $\{\mathcal{M},\mathcal{A}\}^{out} \in \mathbb{R}^{d_{out} \times r }$ and $\{\mathcal{M},\mathcal{A}\}^{in} \in \mathbb{R}^{r \times d_{in}}$.
We can reconstruct the original modulation parameters via matrix-multiplication.
For bias parameter, because of its single dimensionality, we do not apply weight decomposition.
\vspace{-0.3cm}
\section{Determination of Task Sequence}
We constructed a sequence of tasks using perceptual distances between domains.
To measure distance, we adopted Learned Perceptual Image Patch Similarity \cite{zhang2018unreasonable} (LPIPS).
We computed pairwise LPIPS distance between two domain pair.
The results are shown on Table \ref{table:task_sequence}.

\begin{table}[h]
\resizebox{\columnwidth}{!}{
\begin{tabular}{c|ccccc}
\toprule[1.1pt]
 Domains   & Sketches       & Female         & Sunglasses     & Male           & Babies         \\ \hline
FFHQ       & \textbf{0.735} & 0.253          & 0.571          & 0.309          & 0.531          \\
Sketches   &                & \textbf{0.697} & 0.665          & 0.688          & 0.683          \\
Female     &                &                & \textbf{0.523} & 0.266          & 0.480          \\
Sunglasses &                &                &                & \textbf{0.498} & 0.497          \\
Male       &                &                &                &                & \textbf{0.471} \\ \bottomrule[1.1pt]
\end{tabular}
}
\caption{Pairwise LPIPS distance between domains. The large value represents that the two domains are perceptually distant.}
\label{table:task_sequence}
\end{table}
For the most challenging setting, we organized the task order by assigning the least similar task as the next task compared to the current task.
As a result, we were able to decide the sequence of tasks as Sketches $(\mathcal{T}_1)$ , Female $(\mathcal{T}_2)$, Sunglasses $(\mathcal{T}_3)$, Male $(\mathcal{T}_4)$, and Babies $(\mathcal{T}_5)$.
\section{Implementation Detail}
We used StyleGAN2 \cite{karras2020analyzing}\footnote{\url{https://github.com/rosinality/stylegan2-pytorch}} as a backbone of our framework.
Our training configurations came from \cite{ojha2021few, zhao2022closer}.
We adopted Patch Discriminator proposed in \cite{ojha2021few}.
We trained our model using Adam optimizer \cite{kingma2014adam} with the learning rate of 0.002.
The batch size was set to 4.
We experimented on a single GeForce RTX 3090 GPU.
When we train our framework on each task, we froze the pre-trained weights and adopted LeFT modulators on them.
We only saved the lightweight LeFT modulators after learning on each task.
In the inference step, we loaded LeFT modulator to our backbone to generate the images of the previous tasks.
\vspace{-1cm}
\begin{figure}[!]
    \centering
    \includegraphics[width=\columnwidth]{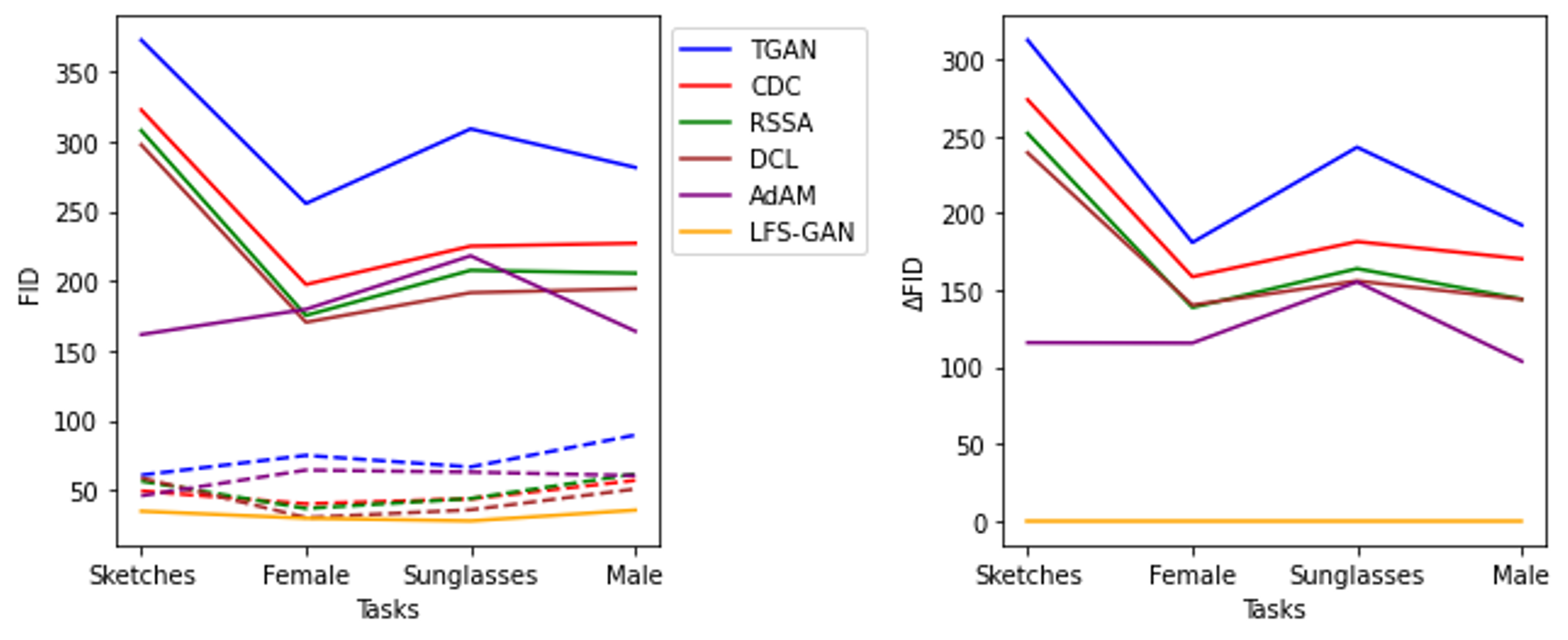}
    \caption{(left): We compared FID \cite{heusel2017gans} scores after training on each tasks and last task in lifelong few-shot image generation task. The solid lines represent the results after the last task, while the dashed lines represent those of each tasks. (right): We visualized the difference of FID scores.}
    \label{fig:forgetting}
\end{figure}
\begin{figure*}[!]
        \centering
        \includegraphics[height=0.45\textheight]{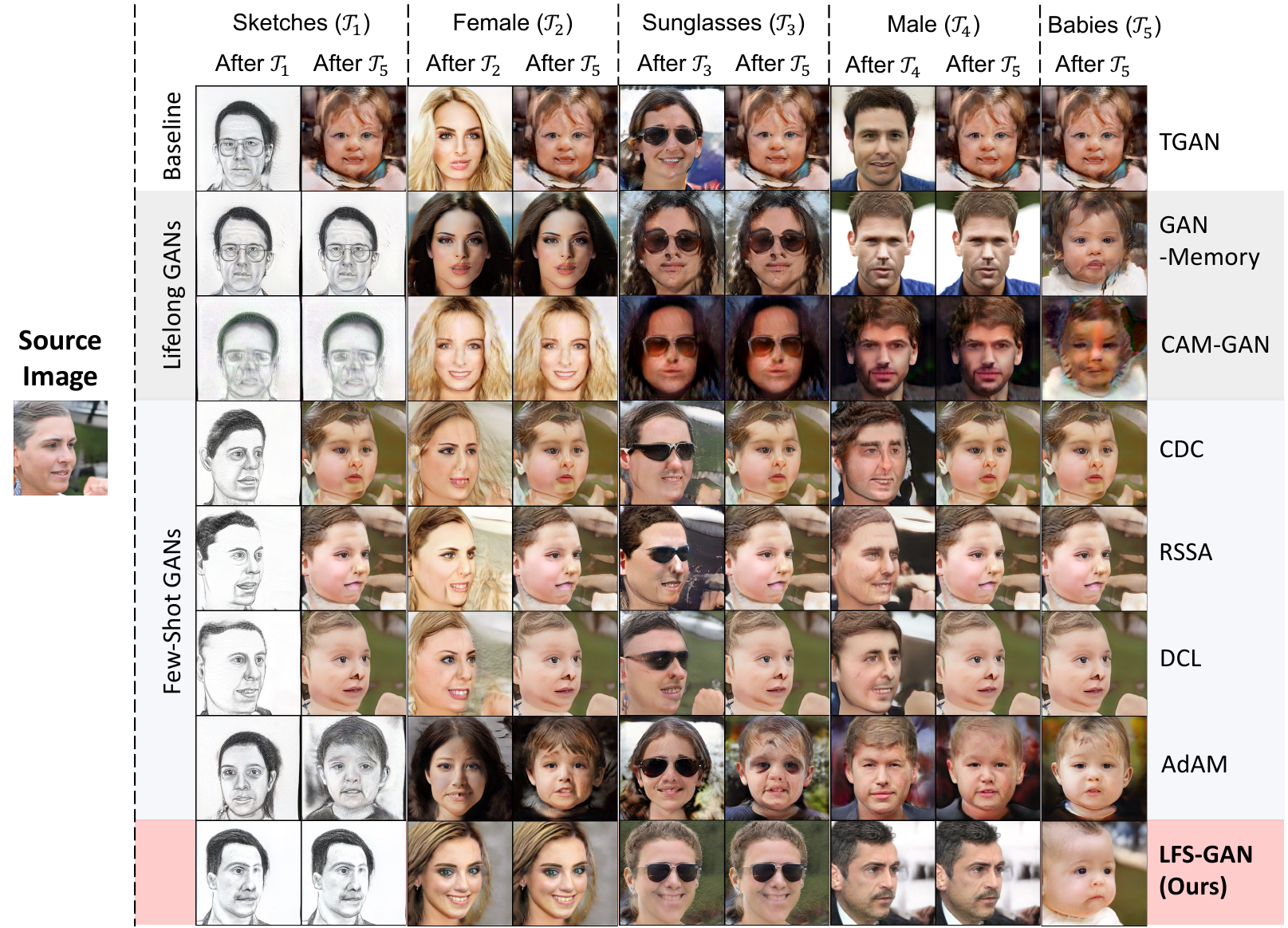}
        \caption{Qualitative results on lifelong few-shot image generation task.}
    \label{fig:lfs_qual_supp}
\end{figure*}
\begin{figure}[h]
    \resizebox{\columnwidth}{!}{
    \centering
    \includegraphics{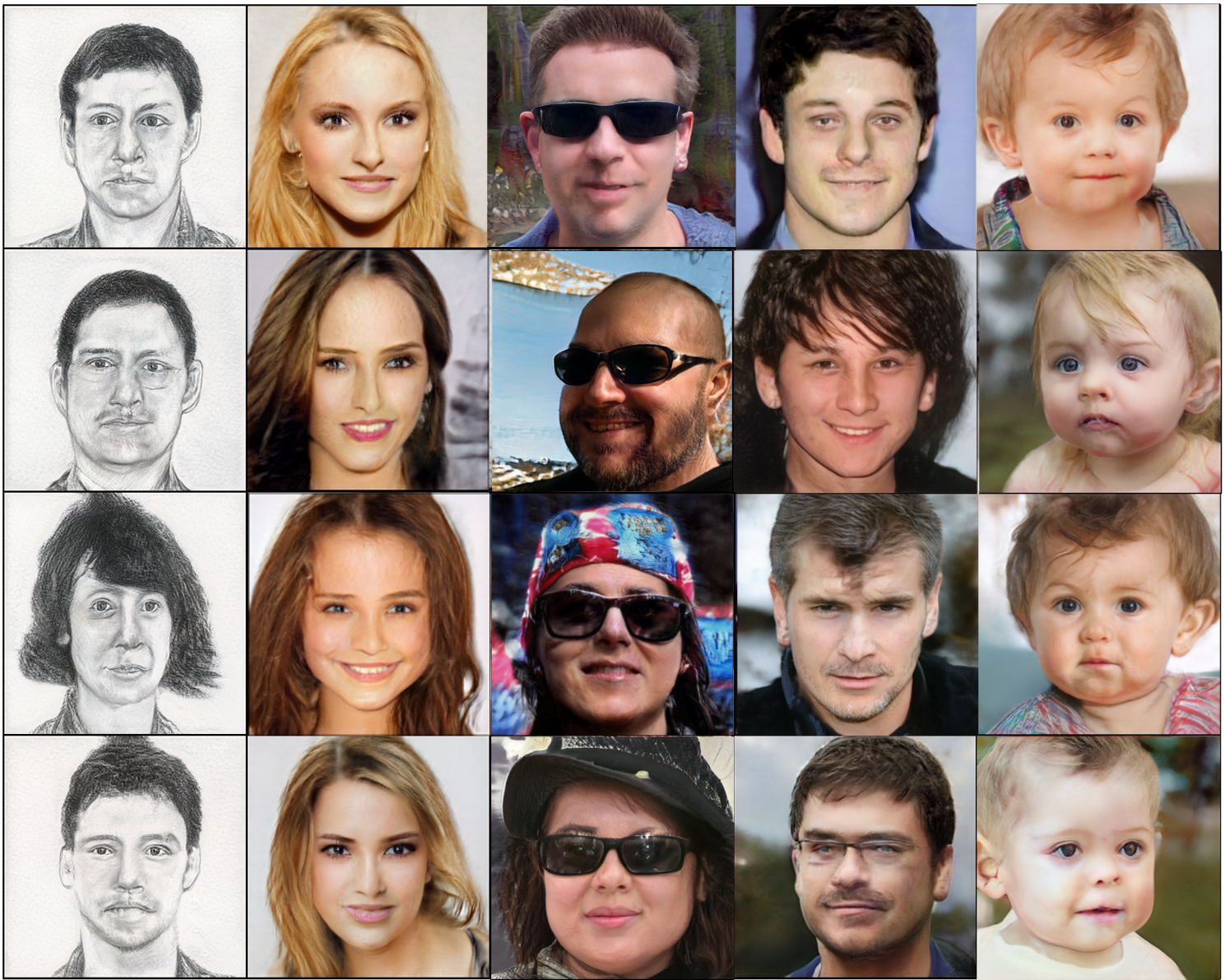}}
    \caption{Diverse samples generated from LFS-GAN. Each column represents each target domain (i.e., Sketches, Female, Sunglasses, Male, and Babies respectively.)}
    \label{fig:diverse_samples}
\end{figure}
\begin{figure}[!]
    \begin{center}
    \resizebox{\columnwidth}{!}{
        \includegraphics{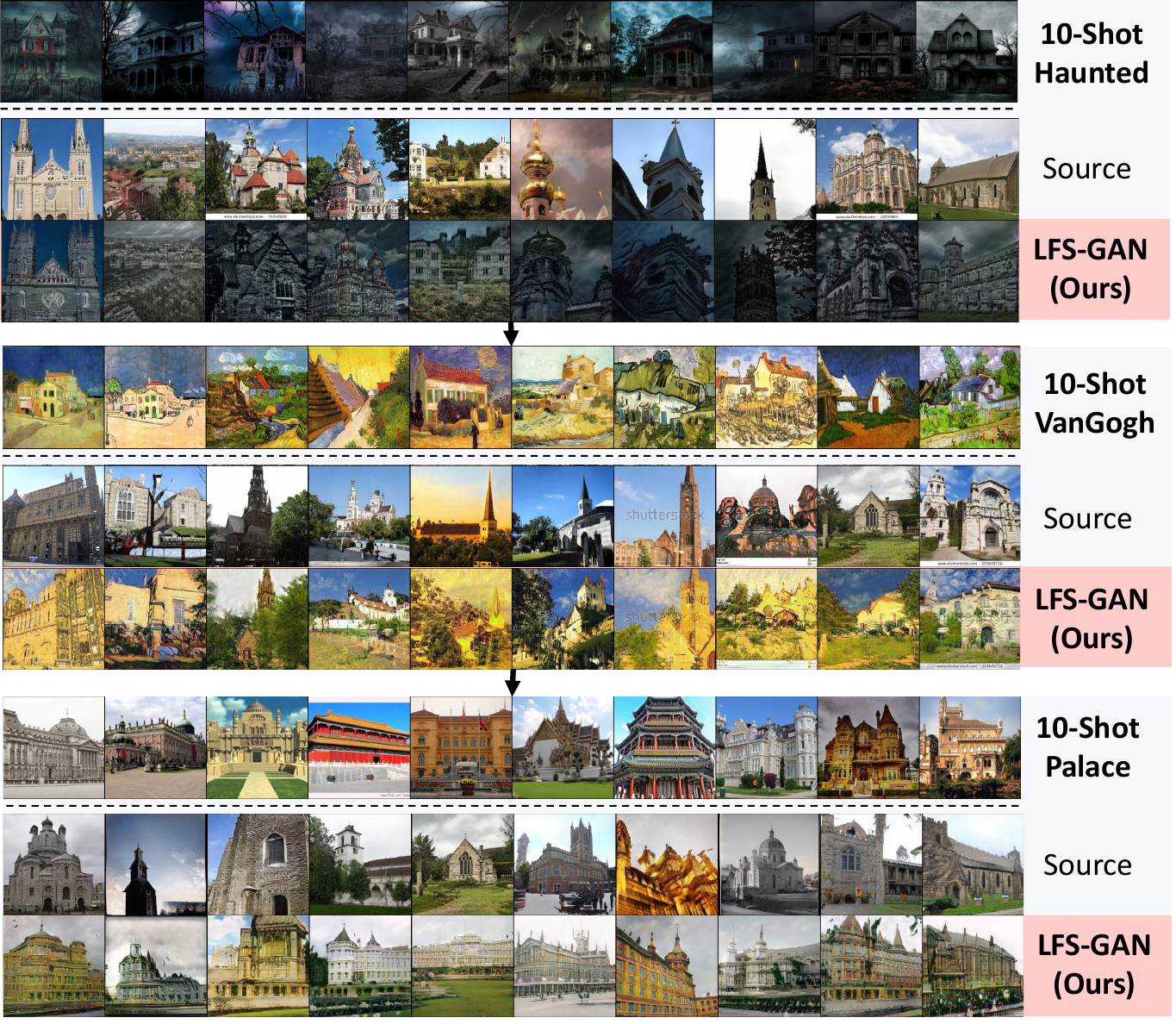}
    }
    \end{center}
    \caption{Qualitative results of our LFS-GAN on lifelong few-shot image generation task. The source domain is LSUN-Church \cite{yu2015lsun}, and the target domains are Haunted houses, Van Gogh's house paintings, and Palace.}
    \label{fig:qual_lfs_church}
\end{figure}
\begin{figure*}[!]
    \begin{center}
    \resizebox{!}{0.4\textheight}{
    \includegraphics{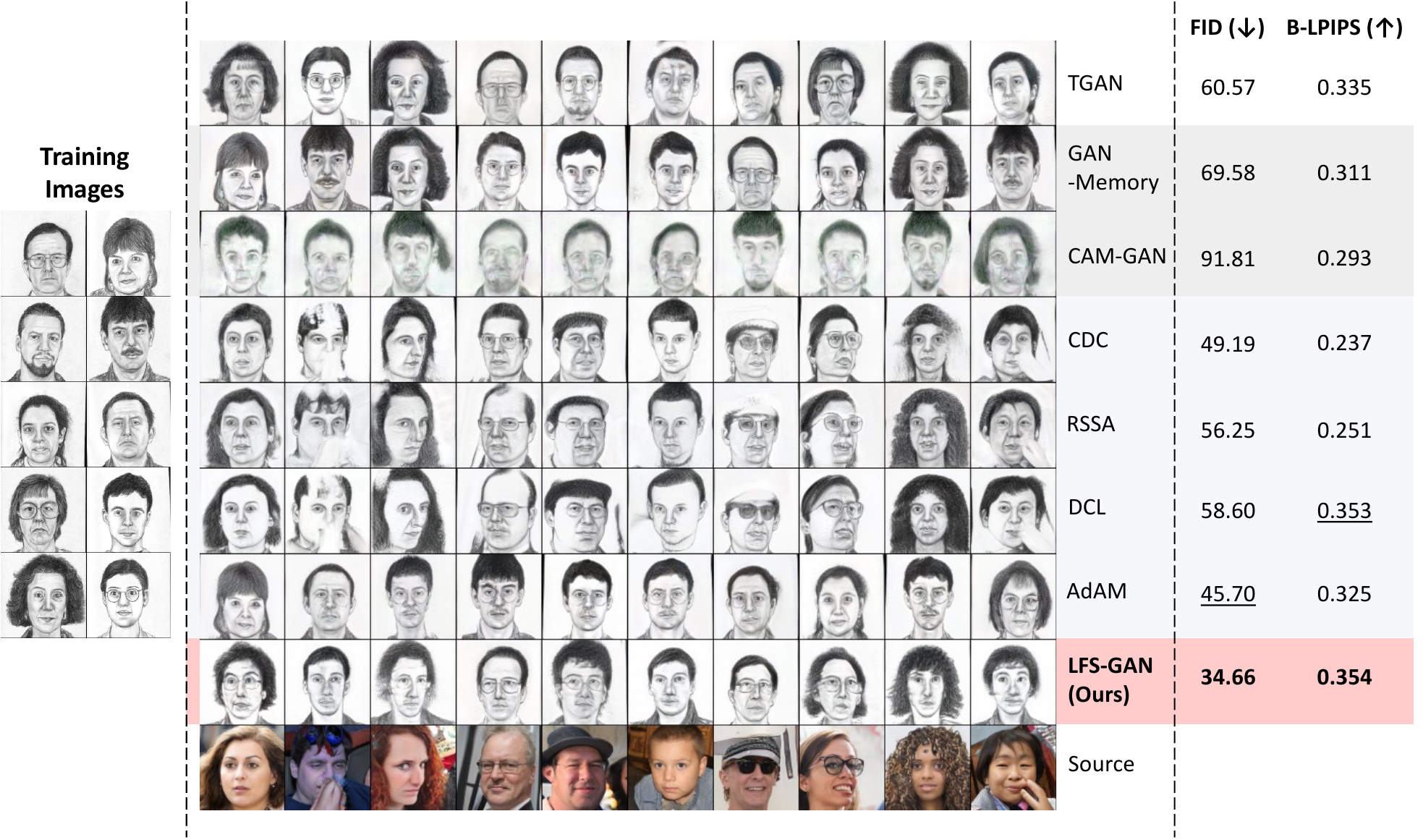}
    }
    \end{center}
    \caption{Qualitative comparison with state-of-the-art lifelong GANs and few-shot GANs on Sketches in few-shot image generation task.}
    \label{fig:qual_fs_sketches}
\end{figure*}
\begin{figure*}[!]
    \begin{center}
    \resizebox{!}{0.4\textheight}{
    \includegraphics{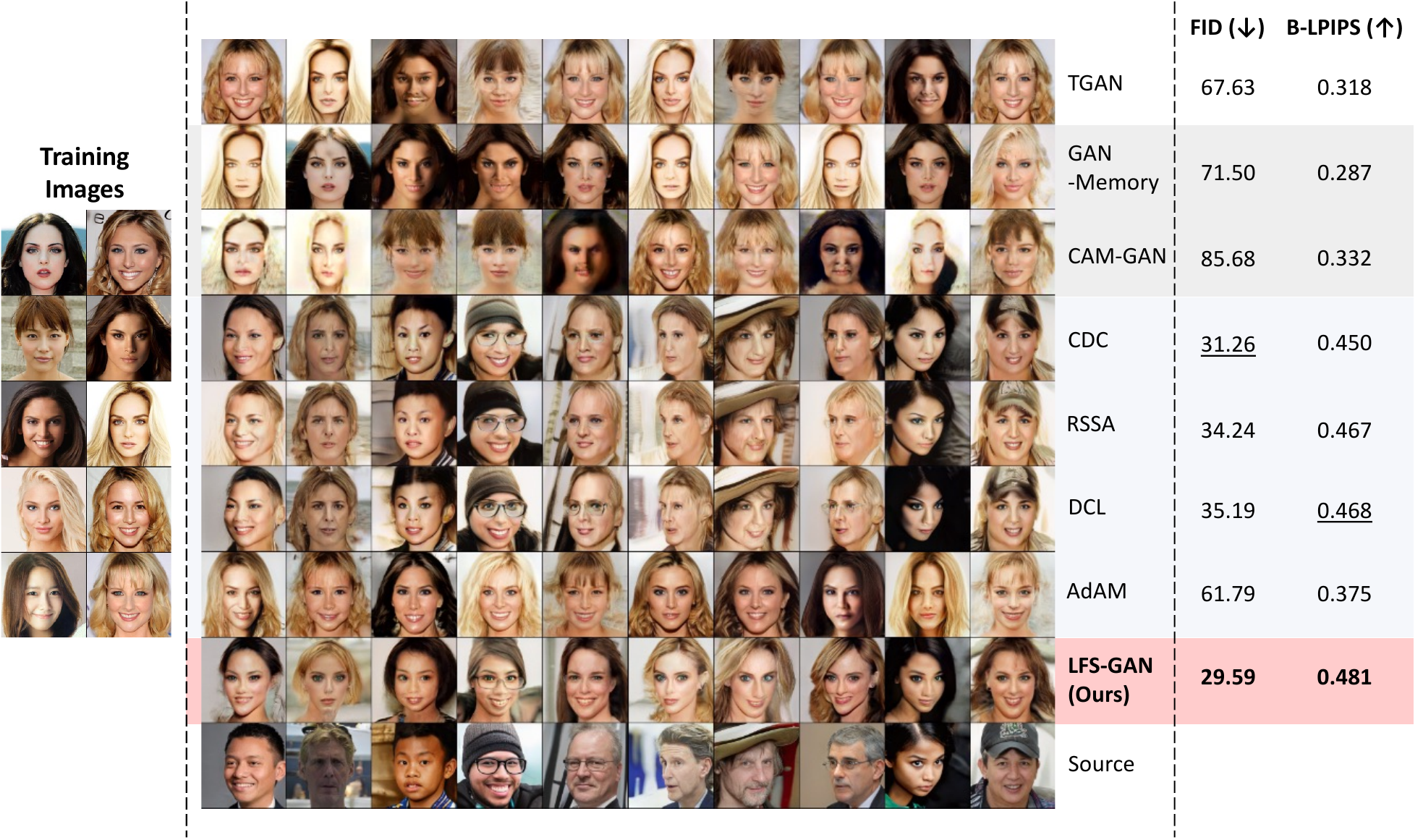}
    }
    \end{center}
    \caption{Qualitative comparison with state-of-the-art lifelong GANs and few-shot GANs on Female in few-shot image generation task.}
    \label{fig:qual_fs_female}
\end{figure*}
\begin{figure*}[!]
    \begin{center}
    \resizebox{!}{0.4\textheight}{
    \includegraphics{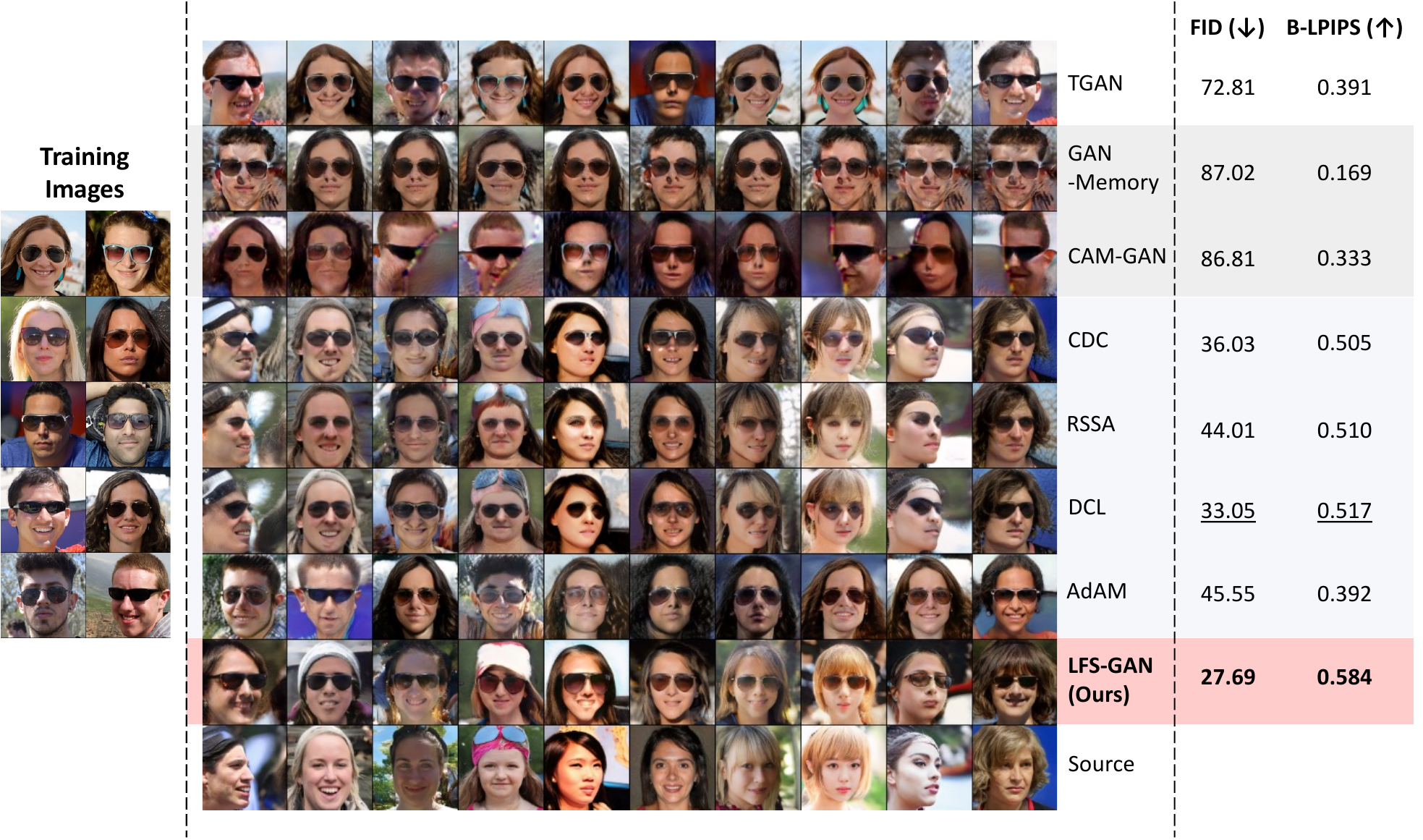}
    }
    \end{center}
    \caption{Qualitative comparison with state-of-the-art lifelong GANs and few-shot GANs on Sunglasses in few-shot image generation task.}
    \label{fig:qual_fs_sunglasses}
\end{figure*}
\begin{figure*}[!]
    \begin{center}
    \resizebox{!}{0.4\textheight}{
    \includegraphics{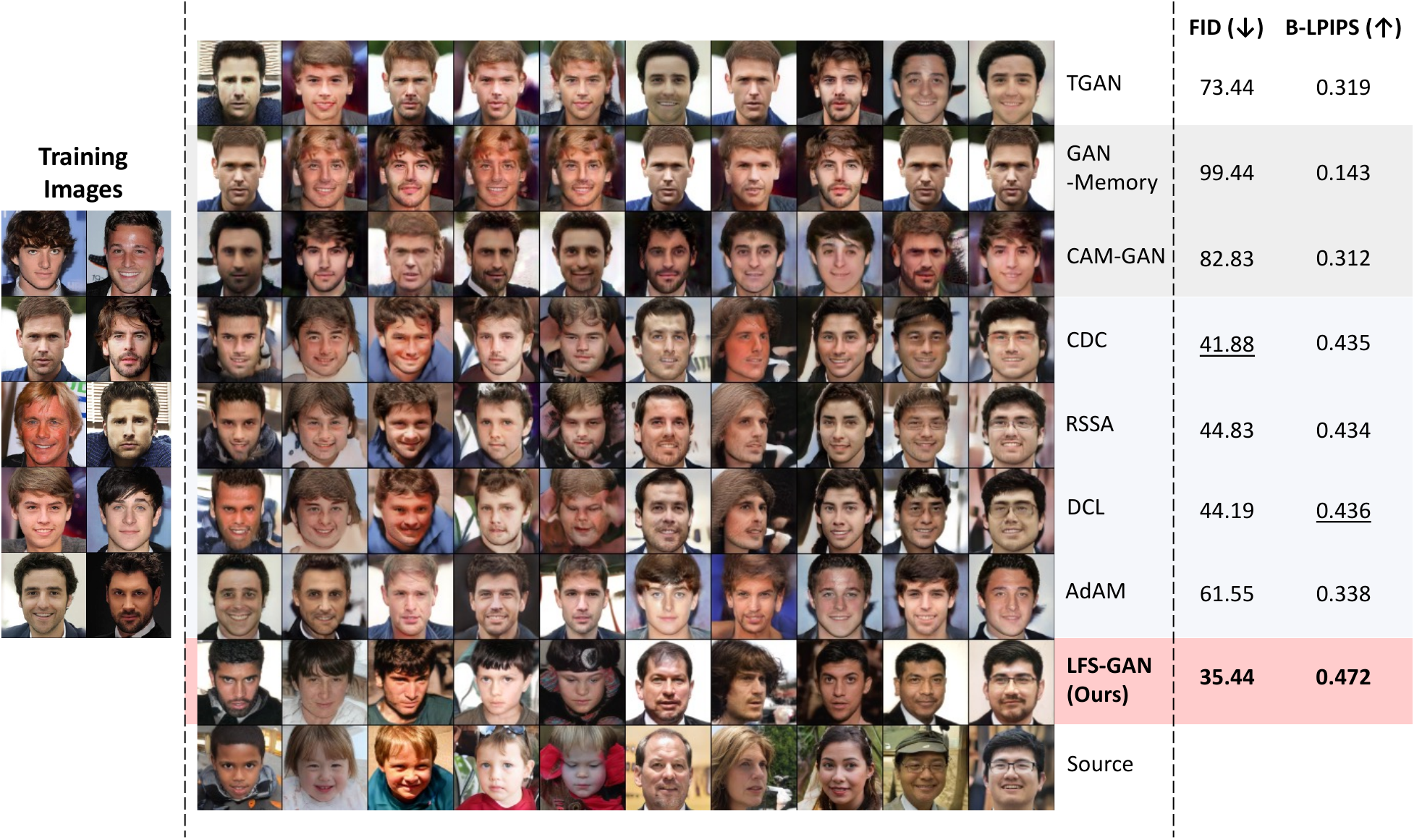}
    }
    \end{center}
    \caption{Qualitative comparison with state-of-the-art lifelong GANs and few-shot GANs on Male in few-shot image generation task.}
    \label{fig:qual_fs_male}
\end{figure*}
\begin{figure*}[!]
    \begin{center}
    \resizebox{!}{0.4\textheight}{
    \includegraphics{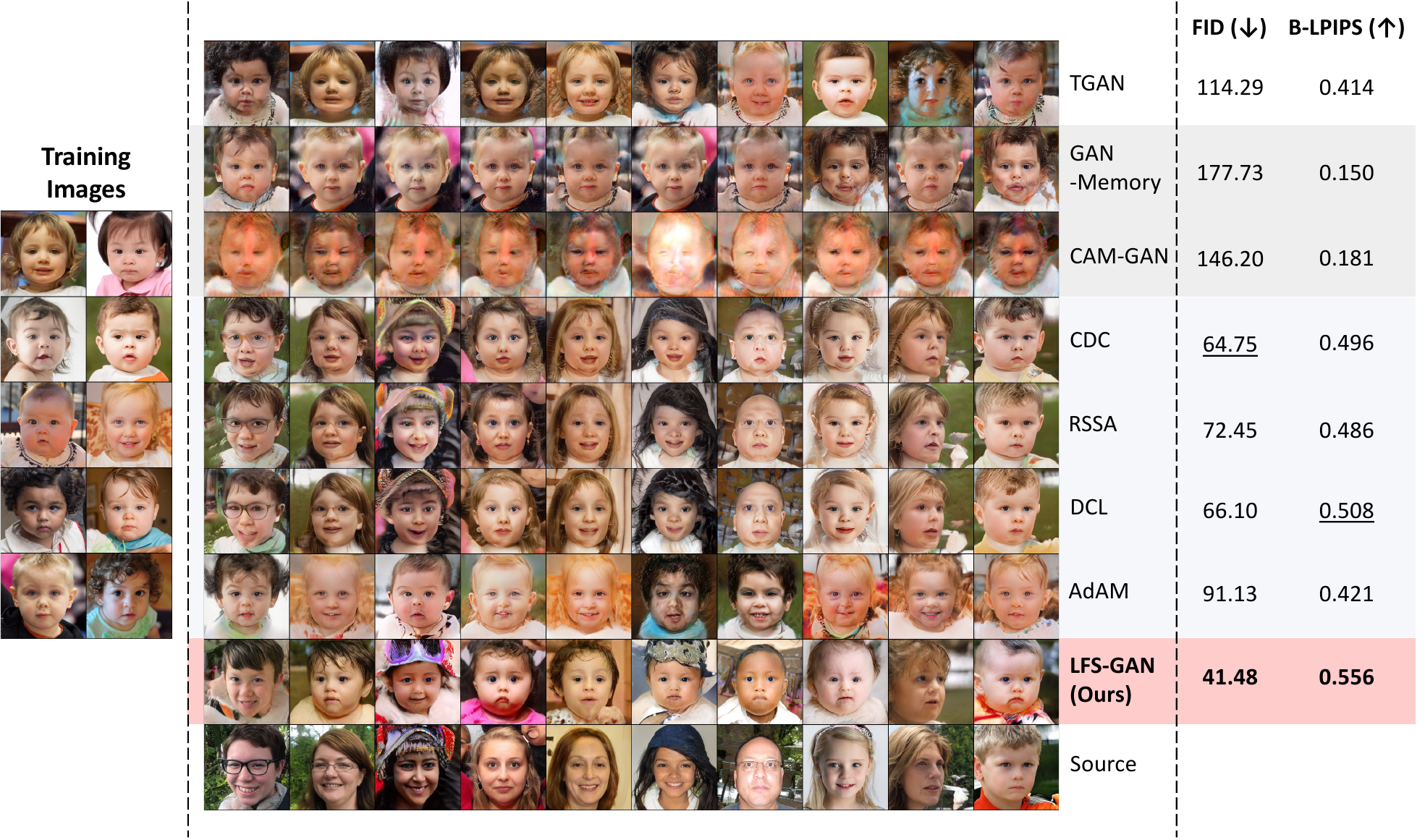}
    }
    \end{center}
    \caption{Qualitative comparison with state-of-the-art lifelong GANs and few-shot GANs on Babies in few-shot image generation task.}
    \label{fig:qual_fs_babies}
\end{figure*}
\begin{figure*}[!]
    \begin{center}
    \resizebox{!}{0.43\textheight}{
    \includegraphics{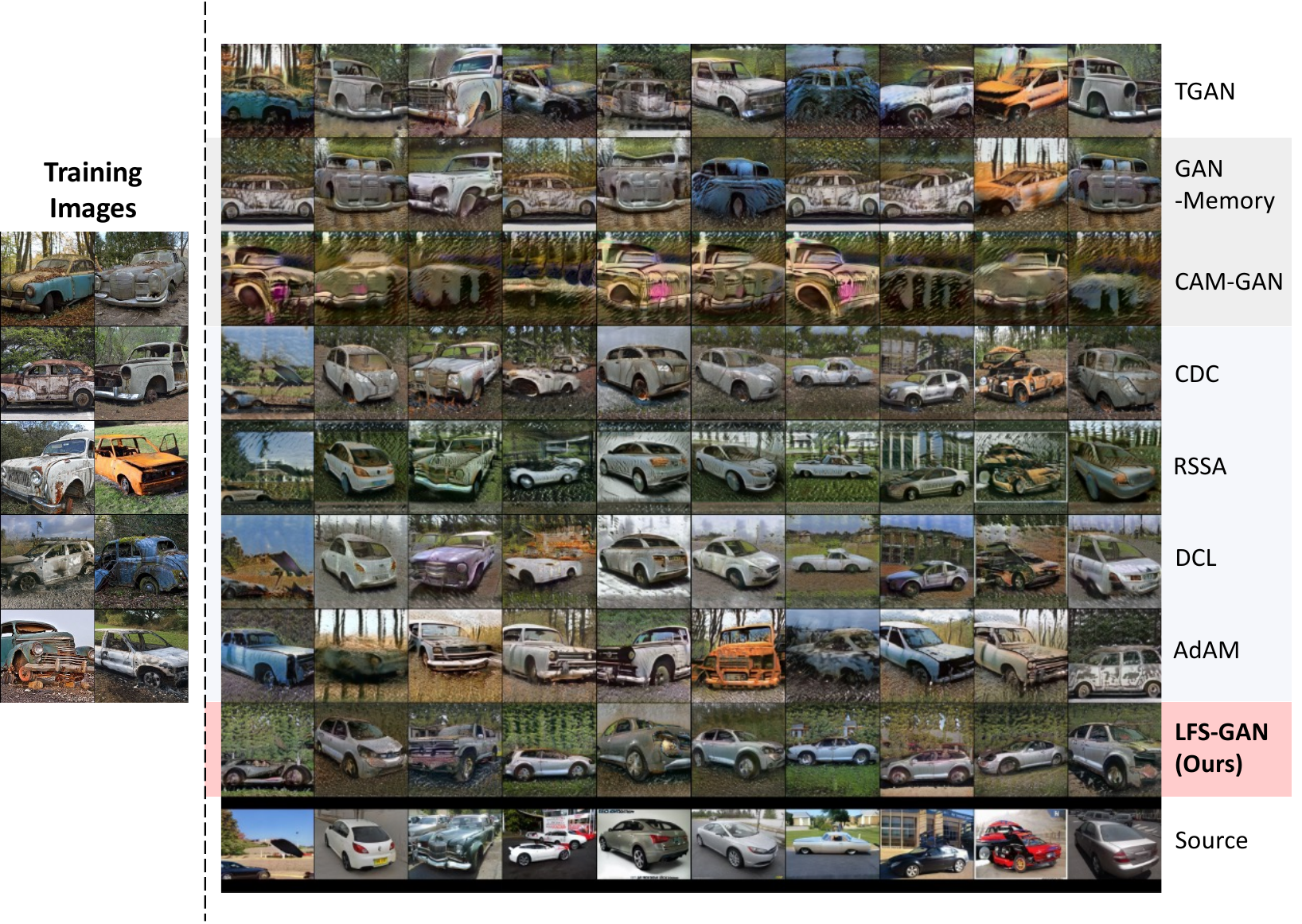}
    }
    \end{center}
    \caption{Qualitative comparison with state-of-the-art lifelong GANs and few-shot GANs on Abandoned cars in few-shot image generation task.}
    \label{fig:qual_fs_wrecked}
\end{figure*}
\begin{figure}[!]
    \centering
    \resizebox{0.8\columnwidth}{!}{
    \includegraphics{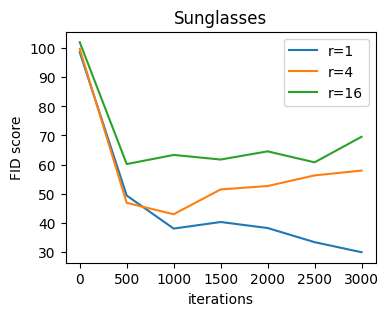}
    }
    \caption{Comparison of FID scores in different ranks.}
    \label{fig:rank_fid}
\end{figure}
\begin{table}[h]
\resizebox{\columnwidth}{!}{%
\begin{tabular}{ccccccccccc}
\toprule[1.1pt]
\multirow{2}{*}{Method} &
  \multicolumn{2}{c}{Sketches ($\mathcal{T}_1$)} &
  \multicolumn{2}{c}{Female ($\mathcal{T}_2$)} &
  \multicolumn{2}{c}{Sunglasses ($\mathcal{T}_3$)} &
  \multicolumn{2}{c}{Male ($\mathcal{T}_4$)} &
  \multicolumn{2}{c}{Babies ($\mathcal{T}_5$)} \\
     & B & I & B & I & B & I & B & I & B & I \\ \hline
AdAM & 0.250   & 0.395   & 0.342   & 0.428   & 0.352   & 0.490   & 0.229   & 0.443   & 0.407   & 0.498   \\
\textbf{LFS-GAN} &
  \textbf{0.354} &
  \textbf{0.405} &
  \textbf{0.481} &
  \textbf{0.546} &
  \textbf{0.584} &
  \textbf{0.631} &
  \textbf{0.472} &
  \textbf{0.561} &
  \textbf{0.556} &
  \textbf{0.627} \\ \bottomrule[1.1pt]
\end{tabular}
}
\caption{Comparison between LFS-GAN and state-of-the-art AdAM on generation diversity in lifelong few-shot image generation task. Here we denote B-LPIPS as `B' and I-LPIPS as `I'.}
\label{tab:tab-ilpips}
\end{table}
\section{Additional Experiments on LFS Task}
\subsection{Comparison on Forgetting}
In Figure \ref{fig:forgetting}, we visualized forgetting occurred in lifelong few-shot image generation task.
We measured FID scores after the each task and after the last task to inspect the performance degradation.
We compared our LFS-GAN with the existing few-shot GANs.
We found that few-shot GANs suffered from catastrophic forgetting, while our LFS-GAN learned multiple tasks without forgetting.
Surprisingly, AdAM \cite{zhao2022fewshot} showed the reduced forgetting compared with other few-shot GANs.
We explain this phenomenon that the weight modulators of AdAM alleviated catastrophic forgetting by recovering a part of previous domain's knowledge.
\subsection{Additional Qualitative Results}
We additionally sampled images from our LFS-GAN and state-of-the-art methods in lifelong few-shot image generation task, the results are shown on Figure \ref{fig:lfs_qual_supp}.
In this figure, which is also shown on Figure 5 of the main paper, lifelong GANs still generated images with a lot of distortions.
On the other hand, few-shot GANs still failed to generate samples of the previous domain $(\mathcal{T}_1\sim\mathcal{T}_4)$.

In Figure \ref{fig:diverse_samples}, we also prepared other images sampled from our LFS-GAN on diverse target domains.
To test our LFS-GAN on different source and target domain pairs, we trained our LFS-GAN from LSUN-Church \cite{yu2015lsun} for a source domain to a sequence of target domains - Haunted houses, Van Gogh's house paintings, and Palace.
The qualitative results are shown on Figure \ref{fig:qual_lfs_church}.
We found that our LFS-GAN could also generate decent images within different source and target domain pairs.
\subsection{Quantitative Comparison using I-LPIPS}
In Table \ref{tab:tab-ilpips}, we compared LFS-GAN with the existing state-of-the-art AdAM on generation diversity.
LFS-GAN outperforms AdAM on both proposed and traditional metrics.
\begin{table*}[h]
\resizebox{\textwidth}{!}{%
\begin{tabular}{cccccccccccc}
\toprule[1.1pt]
\multirow{2}{*}{Method} & \multirow{2}{*}{Setting} &
  \multicolumn{2}{c}{Sketches ($\mathcal{T}_1$)} &
  \multicolumn{2}{c}{Female ($\mathcal{T}_2$)} &
  \multicolumn{2}{c}{Sunglasses ($\mathcal{T}_3$)} &
  \multicolumn{2}{c}{Male ($\mathcal{T}_4$)} &
  \multicolumn{2}{c}{Babies ($\mathcal{T}_5$)} \\
              &  & FID    & B & FID    & B & FID    & B & FID    & B & FID    & B \\ \hline

StyleSwin     & FS                   & 81.79  & 0.25   & 59.29  & 0.31   & 65.23  & 0.34   & 67.95  & 0.34   & 111.46 & 0.34   \\
StyleSwin     & LFS                  & 343.98 & 0.11   & 243.77 & 0.16   & 275.28 & 0.16   & 279.63 & 0.17   & 157.31 & 0.10   \\
StyleSwin+LeFT & LFS                  & 65.48  & 0.31   & 27.31  & 0.50   & 21.89  & 0.54   & 33.94  & 0.41   & 48.81  & 0.51   \\ 
LDM           & FS                   & 112.86 & 0.20   & 105.73 & 0.25   & 85.32  & 0.34   & 106.09 & 0.23   & 144.79 & 0.32   \\
LDM           & LFS                  & 213.09 & 0.03   & 174.48 & 0.06   & 248.94 & 0.12   & 200.17 & 0.13   & 295.15 & 0.05   \\
LDM+LeFT &
  LFS & 75.72 & 0.19   & 73.46  & 0.32   & 71.68 & 0.36   & 94.24  & 0.28   & 132.30 & 0.40
   \\ \bottomrule[1.1pt]
\end{tabular}%
}
\caption{Quantitative results of other methods on few-shot and lifelong few-shot image generation task. We denote B-LPIPS as `B'.}
\label{tab:recent}
\end{table*}
\begin{table}[h]
    \centering
    \begin{tabular}{c|c}
        \toprule[1.1pt]
       Methods  & FID ($\downarrow$) \\ \hline
        StyleGAN2 & 3.62\\
        StyleSwin & 2.81 \\
        LDM & 4.98 \\ \bottomrule[1.1pt]
    \end{tabular}
    \caption{Comparison of generative models on FFHQ 256x256 dataset. To evaluate FID scores, each method samples 50,000 images.}
    \label{tab:backbone_fid}
\end{table}

\begin{table}[h]
\resizebox{\columnwidth}{!}{%
\begin{tabular}{ccccccc}
\toprule[1.1pt]
\multirow{2}{*}{$\lambda$} & \multicolumn{5}{c}{Tasks}                                                          & \multirow{2}{*}{Average} \\
                           & Sketches       & Female         & Sunglasses     & Male           & Babies         &                          \\ \hline
0.25                       & 0.211          & 0.430          & 0.550          & \textbf{0.492} & 0.498          & 0.436                    \\
0.5                        & 0.309          & \underline{0.470}    & \underline{0.558}    & 0.434          & \underline{0.519}    & \underline{0.458}              \\
\textbf{1}                          & \textbf{0.354} & \textbf{0.481} & \textbf{0.584} & 0.472          & \textbf{0.556} & \textbf{0.489}           \\
2                          & 0.220          & 0.431          & 0.512          & \underline{0.489}    & 0.493          & 0.429                    \\
4                          & \underline{0.352}    & 0.448          & 0.510          & 0.453          & 0.504          & 0.453                    \\ \bottomrule[1.1pt]
\end{tabular}%
}
\caption{Ablation on $\lambda$ of the cluster-wise mode seeking loss. We measured B-LPIPS on different $\lambda$.}
\label{tab:ablation_lambda}
\end{table}
\subsection{Experiments on Recent Generative Models}
We tested recent unconditional image generation models - StyleSwin \cite{zhang2022styleswin} and Latent Diffusion Models (LDM) \cite{rombach2022high} on both few-shot and lifelong few-shot image generation tasks.
As shown on Table \ref{tab:recent}, both models also suffered from both catastrophic forgetting and overfitting.
Therefore the challenge of lifelong few-shot image generation task is not limited StyleGAN2 \cite{karras2020analyzing}.
Moreover, when applied LeFT to StyleSwin and LDM, both methods learned few data successfully without any forgetting.
Thus, our proposed LeFT can be generalized on recent generative models.
In this paper, we chose StyleGAN2 as our backbone because it still has been widely utilized and shown comparable performance to other recent generative models like StyleSwin and LDM in FFHQ 256x256 dataset.
We compared generation performances of above architectures in Table \ref{tab:backbone_fid}.
\section{Additional Experiments on FS Task}
\subsection{Qualitative Results}

As shown on Figure \ref{fig:qual_fs_sketches}-\ref{fig:qual_fs_wrecked}, we trained state-of-the-art methods and our LFS-GAN framework on Sketches \cite{ojha2021few}, Female \cite{Karrs2018progressive}, Sunglasses \cite{ojha2021few}, Male \cite{Karrs2018progressive}, and Babies \cite{ojha2021few} from source domain of FFHQ \cite{Karras_2019_CVPR}, and on Abandoned cars from source domain of LSUN-Cars \cite{yu2015lsun}.
In these figures, lifelong GANs generated distorted images and showed a mode collapse problem in all tasks.
Few-shot GANs generated images of better quality.
However, we find that there happened a lot of distortions.
Our LFS-GAN can generate images with reduced distortion and rich diversity compared to the existing state-of-the-art methods.
\begin{figure}[!]
    \centering
    \resizebox{0.9\columnwidth}{!}{
    \includegraphics[height=0.45\textheight]{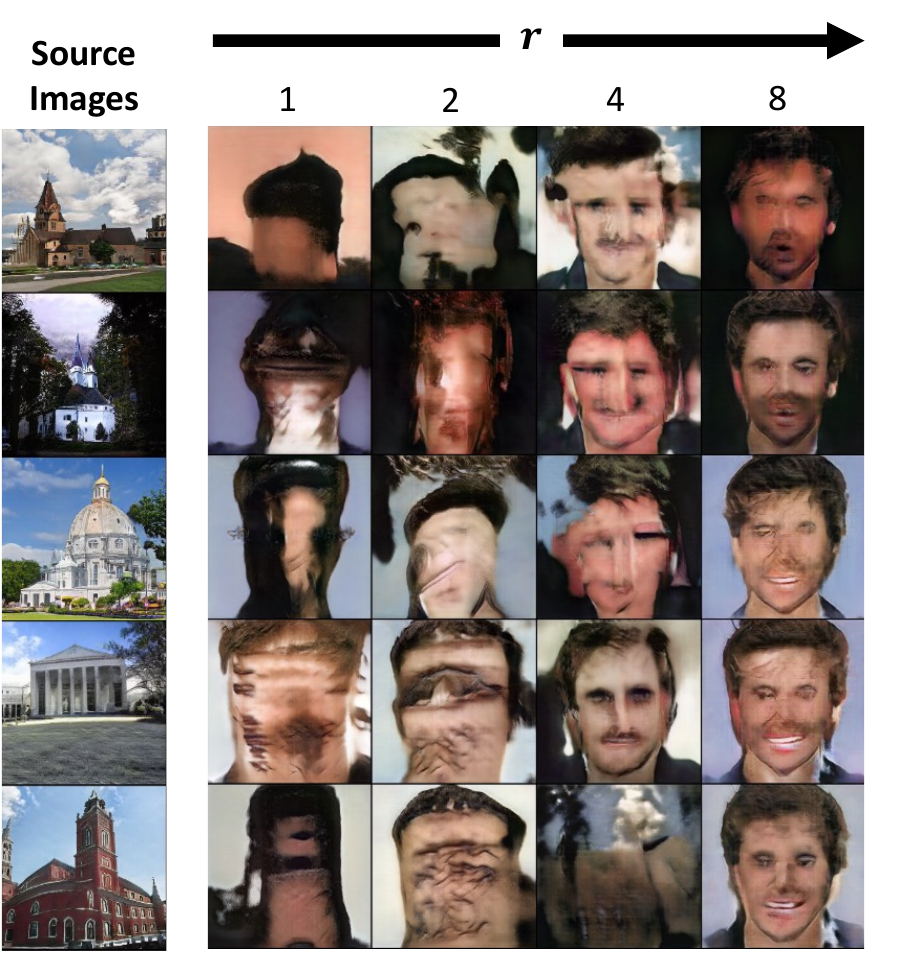}
    }
    \caption{Qualitative results on the distant domain pair (LSUN-Church $\rightarrow$ Male).}
    \label{fig:ablation_r_distant_source}
\end{figure}
\begin{figure}[!]
    \centering
    \resizebox{0.9\columnwidth}{!}{
    \includegraphics[height=0.45\textheight]{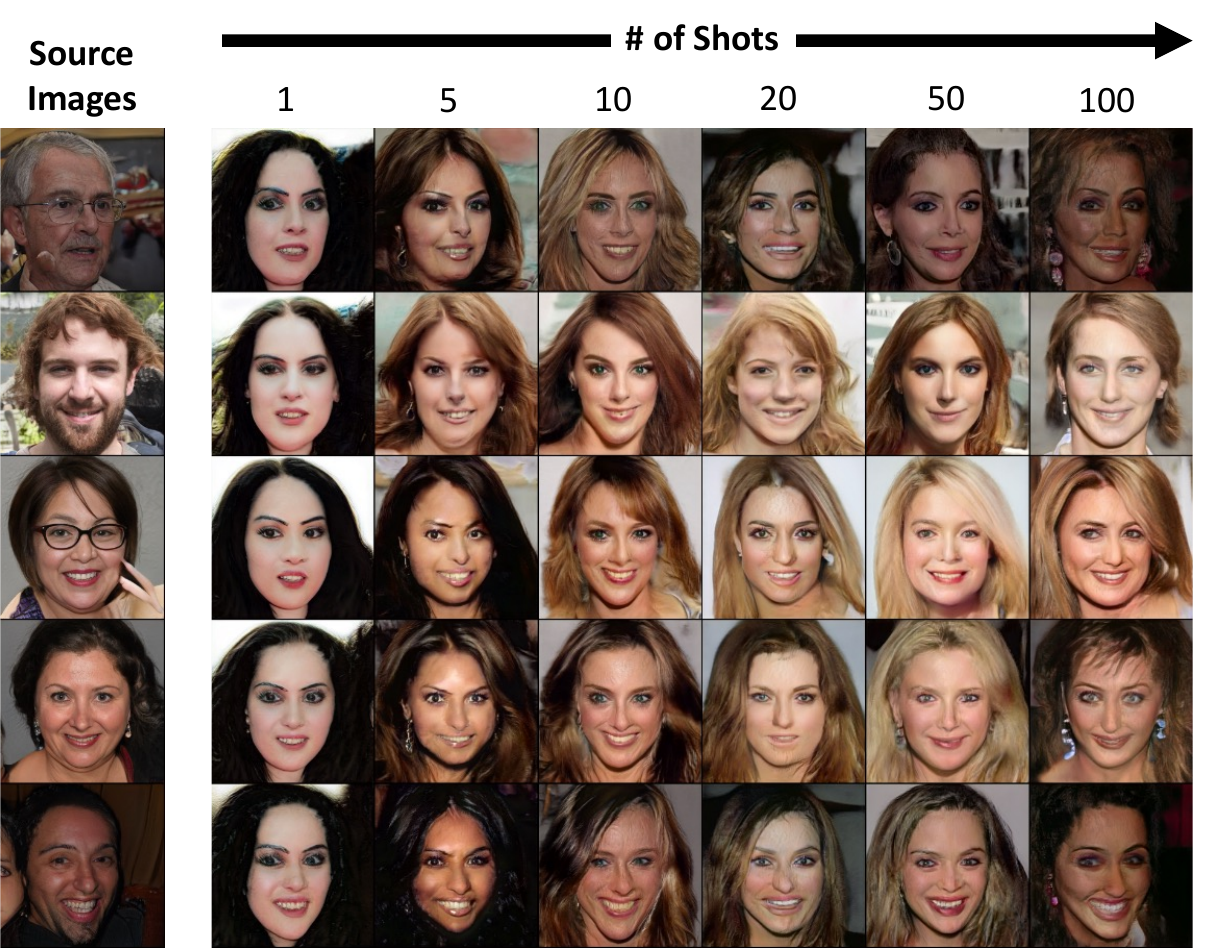}
    }
    \caption{Qualitative results on different number of training images.}
    \label{fig:ablation_shots}
\end{figure}
\begin{table*}[!]
\resizebox{\textwidth}{!}{
\begin{tabular}{ccccccccccccccc}
\toprule[1.1pt]
\multirow{2}{*}{Bias} &
  \multirow{2}{*}{$r$} &
  \multirow{2}{*}{\begin{tabular}[c]{@{}c@{}}\# of\\ Trainable Params.\end{tabular}} &
  \multicolumn{2}{c}{Sketches} &
  \multicolumn{2}{c}{Female} &
  \multicolumn{2}{c}{Sunglasses} &
  \multicolumn{2}{c}{Male} &
  \multicolumn{2}{c}{Babies} &
  \multicolumn{2}{c}{Average} \\ \cline{4-15} 
 &
   &
   &
  FID &
  B-LPIPS &
  FID &
  B-LPIPS &
  FID &
  B-LPIPS &
  FID &
  B-LPIPS &
  FID &
  B-LPIPS &
  FID ($\downarrow$) &
  B-LPIPS ($\uparrow$) \\ \hline
\multirow{5}{*}{\textbf{w/}} &
  \textbf{1} &
  108K &
  \textbf{34.66} &
  \textbf{0.354} &
  \textbf{29.59} &
  \underline{0.481} &
  \textbf{27.69} &
  \textbf{0.584} &
  \textbf{35.44} &
  \textbf{0.472} &
  \textbf{41.48} &
  \textbf{0.556} &
  \textbf{33.77} &
  \textbf{0.489} \\
 &
  2 &
  192K &
  \underline{35.19} &
  0.237 &
  35.96 &
  0.447 &
  34.85 &
  \underline{0.537} &
  43.58 &
  0.417 &
  51.53 &
  0.479 &
  40.22 &
  \underline{0.423} \\
 &
  4 &
  358K &
  38.08 &
  0.248 &
  39.86 &
  0.383 &
  42.92 &
  0.463 &
  44.67 &
  0.422 &
  \underline{49.49} &
  0.493 &
  43.00 &
  0.402 \\
 &
  8 &
  695K &
  35.85 &
  0.163 &
  40.68 &
  0.405 &
  45.74 &
  0.468 &
  54.15 &
  0.331 &
  64.90 &
  0.486 &
  48.27 &
  0.370 \\
 &
  16 &
  1,380K &
  40.74 &
  0.232 &
  48.46 &
  0.304 &
  53.00 &
  0.407 &
  56.14 &
  0.349 &
  77.65 &
  0.367 &
  55.20 &
  0.332 \\ \hline
\multirow{5}{*}{w/o} &
  1 &
  \textbf{54K} &
  37.52 &
  0.223 &
  \underline{34.11} &
  \textbf{0.492} &
  \underline{33.14} &
  0.448 &
  \underline{40.19} &
  \underline{0.423} &
  52.15 &
  0.494 &
  \underline{39.42} &
  0.416 \\
 &
  2 &
  96K &
  35.67 &
  \underline{0.274} &
  34.59 &
  0.404 &
  34.07 &
  0.418 &
  41.46 &
  0.417 &
  52.04 &
  0.536 &
  39.56 &
  0.410 \\
 &
  4 &
  180K &
  41.66 &
  0.256 &
  36.86 &
  0.402 &
  40.87 &
  0.423 &
  42.62 &
  0.370 &
  59.98 &
  \underline{0.556} &
  44.40 &
  0.402 \\
 &
  8 &
  350K &
  41.22 &
  0.201 &
  42.25 &
  0.364 &
  43.94 &
  0.468 &
  53.35 &
  0.314 &
  62.93 &
  0.524 &
  48.74 &
  0.374 \\
 &
  16 &
  704K &
  44.67 &
  0.213 &
  46.88 &
  0.313 &
  46.04 &
  0.401 &
  57.69 &
  0.307 &
  66.82 &
  0.395 &
  52.42 &
  0.326 \\ \bottomrule[1.1pt]
\end{tabular}
}
\caption{Ablation on the bias and the rank of LeFT (detailed).}
\label{tab:ablation_r_bias}
\end{table*}

\begin{table*}[!]
\begin{center}
\resizebox{\textwidth}{!}{
\begin{tabular}{ccccccccccccc}
\toprule[1.1pt]
\multirow{2}{*}{Activation} &
  \multicolumn{2}{c}{Sketches} &
  \multicolumn{2}{c}{Female} &
  \multicolumn{2}{c}{Sunglasses} &
  \multicolumn{2}{c}{Male} &
  \multicolumn{2}{c}{Babies} &
  \multicolumn{2}{c}{Average} \\ \cline{2-13} 
 &
  FID &
  B-LPIPS &
  FID &
  B-LPIPS &
  FID &
  B-LPIPS &
  FID &
  B-LPIPS &
  FID &
  B-LPIPS &
  FID ($\downarrow$)&
  B-LPIPS ($\uparrow$) \\ \hline
Identity &
  45.25 &
  0.285 &
  30.77 &
  \underline{0.475} &
  32.66 &
  0.525 &
  39.64 &
  \underline{0.462} &
  51.00 &
  0.509 &
  39.87 &
  \underline{0.451} \\
Sigmoid &
  34.16 &
  0.250 &
  \textbf{28.92} &
  0.452 &
  31.10 &
  0.496 &
  41.77 &
  0.383 &
  52.85 &
  0.503 &
  37.76 &
  0.417 \\
Tanh &
  40.86 &
  \underline{0.292} &
  30.92 &
  0.458 &
  31.24 &
  0.523 &
  37.97 &
  0.442 &
  50.25 &
  0.486 &
  38.25 &
  0.440 \\
LeakyReLU &
  \textbf{33.43} &
  0.277 &
  49.65 &
  0.451 &
  30.77 &
  0.505 &
  \textbf{30.96} &
  0.449 &
  \textbf{32.39} &
  0.504 &
  \underline{35.42} &
  0.437 \\
GELU &
  41.94 &
  0.322 &
  29.67 &
  0.472 &
  \underline{28.55} &
  \underline{0.541} &
  36.84 &
  0.418 &
  51.88 &
  0.488 &
  37.78 &
  0.448 \\
SiLU &
  41.06 &
  0.251 &
  32.78 &
  0.418 &
  32.82 &
  0.519 &
  36.57 &
  0.390 &
  58.17 &
  \underline{0.510} &
  40.28 &
  0.417 \\
\textbf{ReLU} &
  \underline{34.66} &
  \textbf{0.354} &
  \underline{29.59} &
  \textbf{0.481} &
  \textbf{27.69} &
  \textbf{0.584} &
  \underline{35.44} &
  \textbf{0.472} &
  \underline{41.48} &
  \textbf{0.556} &
  \textbf{33.77} &
  \textbf{0.489} \\ \bottomrule[1.1pt]
\end{tabular}
}
\end{center}
\caption{Ablation on the activation functions of LeFT (detailed).}
\label{tab:ablation_act}
\end{table*}

\begin{table*}[h]
\centering
\resizebox{0.7\textwidth}{!}{
\begin{tabular}{ccccccccccccccc}
\toprule[1.1pt]
\multicolumn{3}{c}{Maximize} & 
  \multirow{2}{*}{Sketches} &
  \multirow{2}{*}{Female} &
  \multirow{2}{*}{Sunglasses} &
  \multirow{2}{*}{Male} &
  \multirow{2}{*}{Babies} &
  \multirow{2}{*}{Average} \\ \cline{1-3}
$\Delta w/ \Delta z$ &
  $ \Delta F/ \Delta w$ &
  $ \Delta I/ \Delta w$ &
  \\ \hline
 &
   &
   &
  0.221 &
  0.427 &
  0.499 &
  0.418 &
  \underline{0.551} &
  0.423 \\ \hline
 &
   &
  \checkmark &
  \underline{0.282} &
  \underline{0.469} &
  0.473 &
  \underline{0.447} &
  0.511 &
  \underline{0.436} \\
 &
  \checkmark &
   &
  0.267 &
  0.450 &
  0.526 &
  0.397 &
  0.486 &
  0.426 \\
\checkmark &
   &
   &
  0.278 &
  0.446 &
  \underline{0.537} &
  0.415 &
  0.500 &
  0.435 \\ \hline
\checkmark &
  \checkmark &
  \checkmark &
  \textbf{0.354} &
  \textbf{0.481} &
  \textbf{0.584} &
  \textbf{0.472} &
  \textbf{0.556} &
  \textbf{0.489} \\ \bottomrule[1.1pt]
\end{tabular}
}
\caption{Ablation on the maximization target of the cluster-wise mode seeking loss (detailed). We measuresd B-LPIPS on different maximization target in our proposed cluster-wise mode seeking loss.}
\label{tab:ablation_cms}
\end{table*}

\section{Additional Ablation Studies}
\subsection{Effect of Rank}
In Figure \ref{fig:rank_fid}, we compared FID scores in different rank.
In the graph, the larger ranks (orange and green lines) tended to perform worse and diverge, while the smallest rank (blue line) consistently achieved convergence.
To test our LFS-GAN in a semantically large domain gap between source and target, we set the source domain as LSUN-Church and the target domains as the existing facial domains.
The results are shown on Figure \ref{fig:ablation_r_distant_source}.
Note that the rank of 8 was the most effective in this large domain gap setting.
\subsection{The Number of Training Images}
In Figure \ref{fig:ablation_shots}, we presented the performance difference according to the number of training images.
We observed that as the number of training images increased, the generation quality also increased.
\subsection{$\lambda$ of Cluster-Wise Mode Seeking Loss}
As shown on Table \ref{tab:ablation_lambda}, we inspected the effect of $\lambda$ of our proposed cluster-wise mode seeking loss.
While we found that $\lambda$=1 was the most effective, our LFS-GAN showed comparable performance on different $\lambda$ values.
\subsection{Detailed Results of Ablation Studies}
We described the detailed results of ablation studies which were previously stated in the main paper on Table \ref{tab:ablation_r_bias}, \ref{tab:ablation_act}, and \ref{tab:ablation_cms}.
We found that using bias, the rank of 1, and ReLU activation generally showed the superior performances compared to other options.
Furthermore, we confirmed that maximizing the relative distance of intermediate latent vectors $(w)$, feature maps $(F)$, and generated images $(I)$ was the most effective for enriching diversity.
\clearpage
{\small
\bibliographystyle{ieee_fullname}
\bibliography{egbib}
}

\end{document}